\documentclass[draftclsnofoot,twocolumn]{IEEEtran}
\usepackage{graphicx}
\usepackage{bm}
\usepackage{subfigure}
\usepackage{epsfig}
\usepackage{latexsym}
\usepackage{amsmath}
\usepackage{amsthm}
\usepackage{amssymb}
\usepackage{amsfonts}
\usepackage{color}
\usepackage{enumerate}
\usepackage{multirow}
\usepackage{booktabs}
\usepackage{epstopdf}% esp to pdf
\usepackage{algorithm}
\usepackage{algpseudocode}
\usepackage{threeparttable}% ±í¸ñ×¢ÊÍ
\usepackage{diagbox}
\usepackage{lettrine}
\usepackage{colortbl}
\usepackage{booktabs}
\usepackage{graphicx}
\usepackage{setspace}
\usepackage[utf8]{inputenc}

\definecolor{mycI1}{rgb}{0,0,0.8125}
\definecolor{mycI2}{rgb}{0,0.0625,1}
\definecolor{mycI3}{rgb}{0,0.3125,1}
\definecolor{mycI4}{rgb}{0,0.5625,1}
\definecolor{mycI5}{rgb}{0,0.8125,1}
\definecolor{mycI6}{rgb}{0.0625,1,0.9375}
\definecolor{mycI7}{rgb}{0.3125,1,0.6875}
\definecolor{mycI8}{rgb}{0.5625,1,0.4375}
\definecolor{mycI9}{rgb}{0.8125,1,0.1875}
\definecolor{mycI10}{rgb}{1,0.9375,0}
\definecolor{mycI11}{rgb}{1,0.6875,0}
\definecolor{mycI12}{rgb}{1,0.4375,0}
\definecolor{mycI13}{rgb}{1,0.1875,0}
\definecolor{mycI14}{rgb}{0.9375,0,0}
\definecolor{mycI15}{rgb}{0.6875,0,0}
\definecolor{mycI16}{rgb}{0.5000,0,0}
\definecolor{mycP1}{rgb}{0,0.1667,1}
\definecolor{mycP2}{rgb}{0,0.5000,1}
\definecolor{mycP3}{rgb}{0,1,1}
\definecolor{mycP4}{rgb}{0.5000,1,0.5000}
\definecolor{mycP5}{rgb}{0.8333,1,0.667}
\definecolor{mycP6}{rgb}{1,0.6667,0}
\definecolor{mycP7}{rgb}{1,0.1667,0}
\definecolor{mycP8}{rgb}{0.8333,0,0}
\definecolor{mycP9}{rgb}{0.5000,0,0}
\definecolor{mycS1}{rgb}{0,0,0.8125}
\definecolor{mycS2}{rgb}{0,0.0625,1}
\definecolor{mycS3}{rgb}{0,0.3125,1}
\definecolor{mycS4}{rgb}{0,0.5625,1}
\definecolor{mycS5}{rgb}{0,0.8125,1}
\definecolor{mycS6}{rgb}{0.0625,1,0.9375}
\definecolor{mycS7}{rgb}{0.3125,1,0.6875}
\definecolor{mycS8}{rgb}{0.5625,1,0.4375}
\definecolor{mycS9}{rgb}{0.8125,1,0.1875}
\definecolor{mycS10}{rgb}{1,0.9375,0}
\definecolor{mycS11}{rgb}{1,0.6875,0}
\definecolor{mycS12}{rgb}{1,0.4375,0}
\definecolor{mycS13}{rgb}{1,0.1875,0}
\definecolor{mycS14}{rgb}{0.9375,0,0}
\definecolor{mycS15}{rgb}{0.6875,0,0}
\definecolor{mycS16}{rgb}{0.5000,0,0}
\definecolor{tabcolor}{rgb}{1,0,0}
\ifCLASSINFOpdf
\else
\fi
\begin{document}
\begin{spacing}{1.0}% µ¥±¶Ðоà

\begin{center}
\LARGE
\textbf{Spatial-Spectral Feature Extraction via Deep ConvLSTM Neural Networks for Hyperspectral Image Classification}

\end{center}

%\maketitle
~\\

\textcircled{c} 2022 IEEE.  Personal use of this material is permitted.  Permission from IEEE must be obtained for all other uses, in any current or future media, including reprinting/republishing this material for advertising or promotional purposes, creating new collective works, for resale or redistribution to servers or lists, or reuse of any copyrighted component of this work in other works.

~\\

\textbf{DOI:}\textcolor[rgb]{0.00,0.00,1.00}{\underline{10.1109/TGRS.2019.2961947.}}

~\\

Wen-Shuai~Hu, Heng-Chao~Li, \emph{Senior Member, IEEE}, Lei~Pan, Wei~Li, \emph{Senior Member, IEEE}, Ran~Tao, \emph{Senior Member, IEEE}, and Qian~Du, \emph{Fellow, IEEE}

\newpage

%\title{Feature Extraction and Classification Based on Spatial-Spectral ConvLSTM Neural Network for Hyperspectral Images}
\title{Spatial-Spectral Feature Extraction via Deep ConvLSTM Neural Networks for Hyperspectral Image Classification}

%Manuscript received April 4, 2019; revised April 30, 2019 and September 24, 2019; accepted December 11, 2019.
\author{
\thanks{Manuscript received July 23, 2019; revised October 20, 2019; accepted December 10, 2019. Date of publication January 15, 2020; date of current version May 21, 2020. This work was supported by the National Natural Science Foundation of China under Grant 61871335. (\emph{Corresponding author: Heng-Chao Li.})}
Wen-Shuai~Hu, %~\IEEEmembership{Member,~IEEE,}
        Heng-Chao~Li, \emph{Senior Member, IEEE}, %~\IEEEmembership{Fellow,~OSA,}
        Lei~Pan, Wei~Li, \emph{Senior Member, IEEE}, Ran~Tao, \emph{Senior Member, IEEE}, and Qian~Du, \emph{Fellow, IEEE} %~\IEEEmembership{Life~Fellow,~IEEE}% <-this % stops a space
\thanks{Wen-Shuai. Hu, Heng-Chao Li, and Lei Pan are with the Sichuan Provincial Key Laboratory of Information Coding and Transmission, Southwest Jiaotong University, Chengdu 610031 China (e-mail: hcli@home.swjtu.edu.cn).}
\thanks{Wei Li and Ran Tao are with the School of Information and Electronics, Beijing Institute of Technology, Beijing 100081 China.}
\thanks{Qian Du is with the Department of Electrical and Computer Engineering, Mississippi State University, Mississippi State, MS 39762 USA.}
}%, e-mail: (see http://www.michaelshell.org/contact.html).}% <-this % stops a space

%\thanks{Manuscript received April 19, 2005; revised January 11, 2007.}


\markboth{IEEE TRANSACTIONS ON GEOSCIENCE AND REMOTE SENSING}
{Shell \MakeLowercase{\textit{et al.}}: Bare Demo of IEEEtran.cls for Journals}
%\markboth{Journal of \LaTeX\ Class Files,~Vol.~6, No.~1, January~2007}%
%{Shell \MakeLowercase{\textit{et al.}}: Bare Demo of IEEEtran.cls for Journals}




\maketitle

\begin{abstract}
%\boldmath
In recent years, deep learning has presented a great advance in hyperspectral image (HSI) classification. Particularly, Long Short-Term Memory (LSTM), as a special deep learning structure, has shown great ability in modeling long-term dependencies in the time dimension of video or the spectral dimension of HSIs. However, the loss of spatial information makes it quite difficult to obtain the better performance. In order to address this problem, two novel deep models are proposed to extract more discriminative spatial-spectral features by exploiting the Convolutional LSTM (ConvLSTM). By taking the data patch in a local sliding window as the input of each memory cell band by band, the 2-D extended architecture of LSTM is considered for building the spatial-spectral ConvLSTM 2-D Neural Network (SSCL2DNN) to model long-range dependencies in the spectral domain. To better preserve the intrinsic structure information of the hyperspectral data, the spatial-spectral ConvLSTM 3-D Neural Network (SSCL3DNN) is proposed by extending LSTM to 3-D version for further improving the classification performance. The experiments, conducted on three commonly used HSI data sets, demonstrate that the proposed deep models have certain competitive advantages and can provide better classification performance than other state-of-the-art approaches.

\end{abstract}


% \fi
\IEEEpeerreviewmaketitle

\section{Introduction}

\lettrine[lines=2]{T}{he} hyperspectral remote sensing image is a 3-D data cube, which integrates the spectral information with the 2-D spatial information of land covers. With the development of remote sensing technology, together with the continuous improvement of hyperspectral sensors, hyperspectral images (HSIs) can provide more opportunities to manage and analyze the information from the Earth's surface \cite{DL2002}-\cite{GC2014}. Correspondingly, HSIs have been widely used in many fields, such as environmental sciences \cite{JM2013}, precision agriculture \cite{XZ2016}, \cite{FM2001}, ecological science \cite{AG2010}, and geological exploration \cite{MF2004}.

As a basic and important research topic, HSI classification has attracted plenty of attentions.
%Generally, it can be divided into unsupervised, semisupervised, and supervised methods depending on the label information involved or not. Compared with unsupervised HSI classification methods, the supervised ones can yield the better classification performance due to the exploitation of the label information, while they also present lower complexity when compared with semisupervised methods.
The support vector machine (SVM) is the most widely used classifier and shows great success in HSI classification \cite{LP2017}. Especially, a composite kernel SVM (SVM-CK) was proposed to exploit the spatial and spectral information simultaneously \cite{GC2006}. Inspired by the successful application on face recognition \cite{JW2009}, sparse representation (SR) \cite{MC2015} has also been applied in HSI classification, which was further improved to exploit the spatial information by Chen \emph{et al.} in \cite{YC2011}, resulting in a joint SR classifier (JSRC). Subsequently, more and more classifiers based on joint sparse model were developed, such as non-local weighted JSRC (NLW-JSRC) \cite{HZ2014}, nearest regularized JSRC (NRJSRC) \cite{CC2016}, and correntropy-based robust JSRC (RJSRC) \cite{JP2017}. %However, due to the redundant information and the limited number of labeled samples, it is difficult for these models to obtain more satisfactory performance.
%due to the redundant information and the limited number of labeled samples, it is very difficult for these methods to play their due role in complex scenarios.

In recent years, deep learning has presented a great advance for feature extraction and classification in the field of computer vision, such as object detection \cite{SR2015}, object tracking \cite{GZ2016}, and behavior recognition in crowd scene \cite{AN2017}, and also shown the effectiveness in HSI classification task \cite{FR2010}. An increasing number of feature extraction and classification methods deep learning-based have been designed for HSIs. In \cite{YC2014}, a classification method based on Stacked AutoEncoder (SAE) was exploited for HSI classification for the first time by extracting spatial-spectral features. Since Hu \emph{et al.} \cite{WH2015} and Chen \emph{et al.} \cite{YC2016} introduced Convolutional Neural Network (CNN) into HSI classification, many new classification models based on CNN have emerged and provided the satisfying performance in HSI classification. %Inspired by SVM-CK \cite{GC2006} and CNN \cite{WH2015}, Mei \emph{et al.} \cite{SM2016} proposed a new model based on CNN by using the mean of local patch of central pixel at each band.
Li \emph{et al.} \cite{WL2017} integrated the pixel-pair model into CNN to further obtain discriminative features. Zhao and Du \cite{WS2016} utilized CNN and the balanced local discriminant embedding to fuse the spatial and spectral information. For the sake of joint learning of the spatial-spectral features, many classification models \cite{ZZ2018}-\cite{WS2018} were built by taking the 3-D data cube as the input and deepening the network to obtain more discriminative features. In order to compute the distance between features of the same classes and that between features of different classes more effectively and quickly, Fang \emph{et al.} \cite{LF2019} integrated the hashing learning into CNN to realize the transformation from high-dimensional features to low-dimensional features. In addition, to overcome the problem of the limitation of available labeled samples, two self-taught learning frameworks based on the multiscale Independent Component Analysis (MICA) and the Stacked Convolutional Autoencoder (SCAE) \cite{RK2017}, semisupervised CNN \cite{BL2017}, and a supervised deep feature extraction method based on Siamese CNN (S-CNN) \cite{BL2018} were proposed. In addition to the single-channel deep network models, there are also some novel works \cite{MEP2018}, \cite{HZ2017} based on multi-channel CNN to improve feature extraction and classification. In particular, Li \emph{et al.} \cite{SL2019} made a systematic review of HSI classification methods, and not only comprehensively classified the existing models based on deep learning, but also presented a review of the strategies that are used to improve the performance when the labeled samples are limited, which is meaningful for the future research.

Obviously, CNN has provided an extremely effective and basic structure for various tasks, and become the core architecture of various models, such as VGG16 \cite{SR2015}, ResNet \cite{ZZ2018}, CapsNets \cite{SS2017}, DenseNet \cite{ME2018}, among which the convolutional layer is the core backbone for feature extraction. To address the gradient vanishing problem caused by deeper and deeper structures \cite{RK2015}, effective feature extraction and classification models \cite{ZZ2018}, \cite{SS2017}, \cite{ME2018} were proposed by promoting the CNN filters to improve the classification performance for HSIs.

% Ò»¶ÎдRNN  ²»Ç£³¶ÈκÎLSTM
In order to analyze sequential data, Recurrent Neural Network (RNN) \cite{AG2009}, as an effective deep learning model, has been widely concerned for modeling long-range dependencies, and also expected to extract features for HSI classification. By using a special activation function and an improved Gated Recurrent Units (GRU), Mou \emph{et al.} \cite{LM2017} built a RNN model for pixel-level spectral classification. Considering that the spatial information has a positive impact on HSI classification performance, Liu \emph{et al.} \cite{BL22018} used multiply RNNs to model long-term dependencies between central and neighborhood pixels. Zhang \emph{et al.} \cite{XZ2018} built a local spatial sequential RNN (LSS-RNN) model, in which the low-level features, such as texture and different morphological profile features, were used to construct LSS features, and then input to RNN to extract high-level features. Based on GRU, Hang \emph{et al.} \cite{RH2019} designed an end-to-end RNN model for extracting the useful information from adjacent and non-adjacent spectral bands, and integrated the convolutional layers into RNN to improve the classification accuracy. In addition, Mou \emph{et al.} \cite{LM2019} used RNN to model the temporal dependency of the outputs of convolutional sub-network for change detection in multispectral imagery.

% ÏÂÃæµ¥¶ÀдLSTM
Long Short-Term Memory (LSTM), as a special RNN structure, has demonstrated its stability and power in modeling long-term dependencies in various studies \cite{SH1997}-\cite{IS2014}. The initially proposed structure of LSTM utilizes the special ``memory cells" instead of logistic or tanh hidden units \cite{SH1997}, and there are three significant gate mechanisms in this kind of structure: input gate, output gate, and forget gate, which are used for implementing information protection, transmission, and control, respectively. Specifically, the input gate is applied to control when the input is allowed to be added to the memory cell, the output gate is designed to decide when the input data has an influence on the output of the memory cell, and the forget gate is mainly used for modeling long-range dependencies. It is the design of this special memory cell with fixed weight and self-connected circular edges that ensures the gradient can pass across many time steps without gradient vanishing or explosion problems \cite{ZC2015}. However, there is an inherent drawback in LSTM, in which the spatial structure information will be lost when unfolding the input to the 1-D from. To better model the spatiotemporal relationships, Shi \emph{et al.} \cite{XS2015} extended this data processing method in LSTM to the convolution operation and proposed Convolutional LSTM (ConvLSTM), which consists of convolution structures in both the input-to-state and state-to-state transitions. Due to the characteristics of LSTM, more attentions have been paid to it, and the most common way is to use LSTM in combination with CNN. Wu \emph{et al.} \cite{HW2017} built a hybrid model for HSI classification, in which the convolutional layers followed by the recurrent layers were used to extract spectrally-contextual features. By utilizing the convolutional recurrent layers, Ru{\ss}wurm \emph{et al.} \cite{MR2018} built an encoder framework for land cover classification. Song \emph{et al.} \cite{AS2018} proposed a recurrent 3D fully convolutional network for change detection in HSIs, in which the ConvLSTM layer is used to perform long-term dependencies modeling on the outputs of fully convolutional network in the temporal field. Moreover, Seydgar \emph{et al.} \cite{MS2019} proposed a two-stage model, in which ConvLSTM, cascading with 3D CNN, was utilized to extract spatial-spectral features.

In addition to the above combination with CNN, there are some works to build deep models using LSTM alone. Zhou \emph{et al.} \cite{FZ2017} designed a spatial-spectral LSTMs (SSLSTMs) model based on two independent LSTMs in the way of spatial LSTM (SaLSTM) and spectral LSTM (SeLSTM). However, this way not only insufficiently fuses the spatial and spectral information, but also loses the spatial structure. Ru{\ss}wurm \emph{et al.} \cite{MR2017} employed multi-level cascaded LSTMs for land cover classification. Inspired by spatial similarity measurements \cite{BR2016}, Ma \emph{et al.} \cite{AM2019} built a LSTM-based model, in which the spatial-spectral features are extracted by these measurement strategies. Moreover, a Bidirectional-ConvLSTM (Bi-CLSTM) model \cite{QL2017} was proposed by cascading all outputs in each layer for extracting spatial-spectral features. However, this simple cascade fails to fully utilize the correlation between different spectral bands.

Inspired by the above works, the main purpose of this paper is to construct two novel deep ConvLSTM neural networks for HSI feature extraction and classification. The main contributions of this paper are listed as follows:

\hangindent 2.5em
(1) In order to address the problem of underutilization of the correlations between different spectral bands in Bi-CLSTM and the issues of insufficient fusion of the spatial and spectral information and the loss of the spatial structure information in SSLSTMs, an effective feature extraction model, i.e., spatial-spectral ConvLSTM 2-D Neural Network (SSCL2DNN), is proposed by modeling long-term dependencies in the spectral field for joint learning of spatial-spectral features, in which the local window patch is decomposed into a spectral sequence and then input to each memory cell band by band.

\hangindent 2.5em
(2) To better preserve the intrinsic structure of hyperspectral data, the 3-D structure (namely ConvLSTM3D) is further developed from the basic ConvLSTM cell, with which a novel deep model, i.e., spatial-spectral ConvLSTM 3-D Neural Network (SSCL3DNN), is constructed. Different from the way of band-by-band processing, the local patch directly takes the form of 3D cube as the input of SSCL3DNN, which enables it to further improve classification performance of SSCL2DNN.

The remainder of this paper is organized as follows. Section II presents CNN, LSTM, and ConvLSTM, and discusses the applications on HSI feature extraction and classification. In Sections III and IV, the proposed deep models with the extraction of spatial-spectral features for HSI classification are described in detail, respectively. Comprehensive quantitative analysis and evaluation of the proposed models are implemented in Section V. And conclusively, Section VI summarizes this paper.
\section{Related Work}

\label{sec:relate}
\setlength{\parindent}{1em}
%In this section, we first provide an overview of CNN that has a close connection with the deep models constructed in this paper. Then, a brief introduction for LSTM and its extended version is given, which also is the core backbone of our proposed deep models.
\subsection*{A. Convolutional Neural Network}
The fundamental CNN mainly consists of the following parts: convolutional layer, pooling layer, full connection layer, and classification layer. Based on different convolution operations, we can construct various networks to meet a variety of practical requirements. %, such as speech recognition [58], text recognition [59], object detection and recognition [21], video behavior recognition [23], face detection [60] in computer vision. %Specially, in recent years, with the application and development of deep learning in computer vision, more and more HSI classification methods based on deep learning, especially CNN, have also been greatly developed and achieved very high classification performance.
%Based on three different convolution calculation methods, CNN is composed of three basic forms: 1-D CNN with 1-D convolution operation, 2-D CNN realized by 2-D convolution operation and 3-D CNN with 3-D convolution operation. And the illustrations of 2-D and 3-D convolution operations can be demonstrated in Fig. 1. According to the different convolution operations, we can construct various networks to realize a variety of practical application requirements, such as speech recognition [58], text recognition [59], object detection and recognition [21], video behavior recognition [23], face detection [60] in computer vision. %Specially, in recent years, with the application and development of deep learning in computer vision, more and more HSI classification methods based on deep learning, especially CNN, have also been greatly developed and achieved very high classification performance.
Originally, the calculation formula of each convolutional layer in CNN can be expressed as
\begin{equation}
o_{r+1}=\phi(W_{r}*o_{r}+b_{r}),
\end{equation}
where \(o_{r+1}\) is the output with \({M}\) feature maps of the \(r\)th convolutional layer, \(W_{r}\) denotes the convolution filter, and \(b_{r}\) is the bias of the \(r\)th convolutional layer, \(o_{r}\) is the output of the \((r-1)\)th convolutional layer, and \(\phi(\cdot)\) is the nonlinear activation function. \(W_{r}\) has a size of \(k^{(r)} \times k^{(r)}\) for 2-D CNN. With respect to 3-D CNN, \(W_{r}\) is a convolution filter in the \(r\)th convolutional layer with the size of \(k^{(r)} \times k^{(r)} \times d^{(r)}\), in which \(k^{(r)}\) and \(d^{(r)}\) denote the size and depth of the convolution filter, respectively.
%\begin{figure*}[!htbp]% ÍøÂçÄ£ÐÍ
%\centering
%\vspace{-0.2cm}  %µ÷ÕûͼƬÓëÉÏÎĵĴ¹Ö±¾àÀë
%\setlength{\abovecaptionskip}{0pt}
%\begin{center}
%\includegraphics[width=6.5in]{CNN.jpg}
%\end{center}
%\centering
%\caption{The common network structure of CNN.}
%\vspace{-0.9cm}
%\end{figure*}

Since Hu \emph{et al.} \cite{WH2015} and Chen \emph{et al.} \cite{YC2016} proposed HSI feature extraction and classification models based on CNN, more and more improved CNN-based models have been introduced. %It is undeniable that CNN has provided a extremely effective basic structure for various tasks and increasingly became the core architecture of various models, such as VGG16 [61], ResNet [41],[62], DenseNet [43],[63], CapsNets [42] in which the convolution layer is the core backbone for feature extraction. Therefore, the number of the feature maps output by the convolution layer, the size of the convolution kernel and the number of the convolution layer have a great influence on the performance of the algorithm.
CNN has become an extremely effective and basic structure for various tasks. It is noteworthy that a sliding window is still used to extract spatial features in CNN, which is a traditional way of exploiting spatial information. Moreover, the data transmission in CNN only exists between adjacent layers, lacking information interaction inside each layer, which may make it difficult to extract more effective features.

\subsection*{B. Long Short-Term Memory}
LSTM is proposed to deal with the issues that RNN is not suitable for learning long-term dependencies and prone to bring about gradient vanishing and exploding problems \cite{ZC2015}. It is obvious that the data transmission and processing in LSTM are realized by three key gate units: input gate, output gate, and forget gate, which are used for implementing information protection and control \cite{SH1997}. The calculation formulas between these three gate structures in LSTM are written as
%\vspace{-0.2cm}  %µ÷ÕûͼƬÓëÉÏÎĵĴ¹Ö±¾àÀë
\begin{IEEEeqnarray}{rCl}
\setlength{\abovedisplayskip}{4pt}
\setlength{\belowdisplayskip}{4pt}
i_t &=& \sigma(W_{xi}x_t+W_{hi}h_{t-1}+W_{ci} \circ c_{t-1}+b_i) \notag \\
f_t &=& \sigma(W_{xf}x_t+W_{hf}h_{t-1}+W_{cf} \circ c_{t-1}+b_f) \notag \\
\tilde{c_t} &=& \tanh(W_{xc}x_t+W_{hc}h_{t-1}+b_c) \notag \\
c_t &=& f_t \circ c_{t-1} + i_t \circ \tilde{c_t} \notag \\
o_t &=& \sigma(W_{xo}x_t+W_{ho}h_{t-1}+W_{co} \circ c_t+b_o) \notag \\
h_t &=& o_t \circ \tanh(c_t),
\end{IEEEeqnarray}
where \(x_t\), \(h_{t-1}\), and \(c_{t-1}\) represent the input of the current cell, the output and state of the last cell in LSTM, respectively. \(i_t\), \(f_{t}\), and \(o_{t}\) denote the input gate, forget gate, and output gate of LSTM. \(W_{\bullet i}\) and \(b_{i}\) are the weight and bias of the input gate, \(W_{\bullet f}\) and \(b_{f}\) are the weight and bias of the forget gate, and \(W_{\bullet o}\) and \(b_{o}\) are the weight and bias of the output gate, where \(\bullet\) indicates \(x\), \(h\), and \(c\). \(\circ\) denotes the Hadamard product. \(\sigma\) is the nonlinear activation function.

Zhou \emph{et al.} \cite{FZ2017} first attempted to apply LSTM to HSI classification and advanced the spatial-spectral LSTMs (SSLSTMs). The experimental results show that LSTM can also be well used for modeling long-range dependencies in spectral domain. Nevertheless, it is worth noting that the two branches in SSLSTMs are independent of each other. In addition, the unfolding of the original HSI data to one-dimensional vectors as the input of SSLSTMs will actually lose the intrinsic structure of the hyperspectral data, since spatial information is not considered in LSTM.

\subsection*{C. Convolutional LSTM}
Considering the shortcomings of LSTM, a modification and extended version of LSTM, i.e., ConvLSTM \cite{XS2015}, is developed. Different from LSTM, the input-to-state and state-to-state transitions in ConvLSTM are realized by convolution, and ConvLSTM holds the same structure as LSTM and can be used to model long-term dependencies in the time domain or spectral domain, in which the calculation formulas can be expressed as
\begin{IEEEeqnarray}{rCl}
i_t &=& \sigma(W_{xi}*X_t+W_{hi}*H_{t-1}+W_{ci} \circ C_{t-1}+b_i) \notag \\
f_t &=& \sigma(W_{xf}*X_t+W_{hf}*H_{t-1}+W_{cf} \circ C_{t-1}+b_f) \notag \\
\tilde{C_t} &=&\tanh(W_{xc}*X_t+W_{hc}*H_{t-1}+b_c)  \notag\\
C_t &=& f_t \circ C_{t-1} + i_t \circ \tilde{C_t} \notag \\
o_t &=& \sigma(W_{xo}*X_t+W_{ho}*H_{t-1}+W_{co} \circ C_t+b_o) \notag \\
H_t &=& o_t \circ \tanh(C_t),
\end{IEEEeqnarray}
where \(X_t\) denotes the input of the current cell, \(C_{t-1}\) and \(H_{t-1}\) are state and output of the last cell, respectively. * means the convolution operation. \(W\) denotes the 2-D convolution filter with a \(k \times k\) kernel, and \(k\) is the size of the convolution kernel, respectively. Specially, the definitions of \(W_{\bullet i}\), \(W_{\bullet f}\), \(W_{\bullet o}\), \(b_{i}\), \(b_{f}\), and \(b_{o}\) are similar to that in (2), however, the data dimensions and processing methods are different.
%\(W_{\bullet i}\) and \(b_{i}\) are the weight and bias of the input gate, \(W_{\bullet f}\) and \(b_{f}\) are the weight and bias of the forget gate, and \(W_{\bullet o}\) and \(b_{o}\) are the weight and bias of the output gate, where \(\bullet\) indicates \(x\), \(h\), and \(c\).
%\begin{IEEEeqnarray}{rCl}
%i_t &=& \sigma(W_{xi}*X_t+W_{hi}*H_{t-1}+b_i) \notag \\
%f_t &=& \sigma(W_{xf}*X_t+W_{hf}*H_{t-1}+b_f) \notag \\
%\tilde{C_t} &=&\tanh(W_{xc}*X_t+W_{hc}*H_{t-1}+b_c)  \notag\\
%C_t &=& f_t \circ C_{t-1} + i_t \circ \tilde{C_t} \notag \\
%o_t &=& \sigma(W_{xo}*X_t+W_{ho}*H_{t-1}+b_o) \notag \\
%H_t &=& o_t \circ \tanh(C_t)
%\end{IEEEeqnarray}

%Inspired by this, a HSI feature extraction and classification model based on Bidirectional-Convolutional LSTM in [53] was proposed to extract the spatial-spectral features simultaneously, which solved the problems arising from LSTM and realized a good classification performance for HSI.

\begin{figure*}[htbp]
\centering
%\vspace{-0.2cm}  %µ÷ÕûͼƬÓëÉÏÎĵĴ¹Ö±¾àÀë
\setlength{\abovecaptionskip}{-5pt}
\begin{center}
\includegraphics[height=1.85in, width=6.6in]{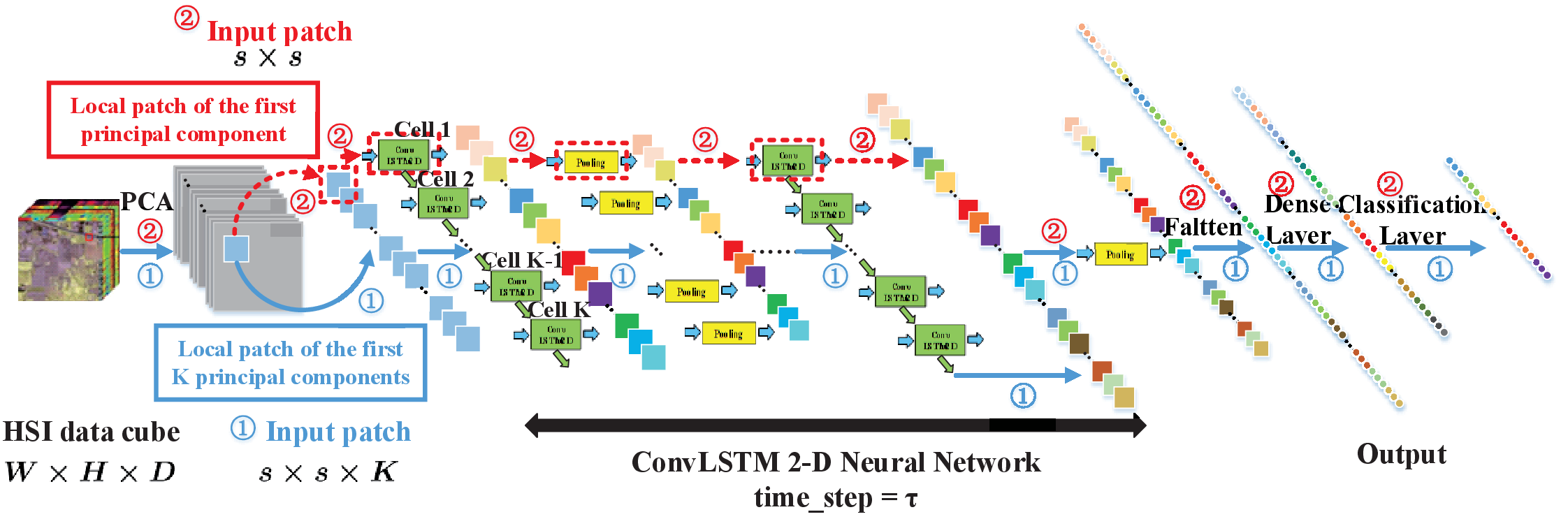}
%\caption{fig3}
\end{center}
%\tiny
\centering
\caption{The proposed spatial-spectral ConvLSTM 2-D Neural Network (SSCL2DNN) is labeled by \(\textcircled{1}\) with blue. In particular, SSCL2DNN can be transformed into another deep model when \(K\) equals 1 and \(\tau\) equals 1, which is called spatial ConvLSTM 2-D Neural Network (SaCL2DNN) labeled by \(\textcircled{2}\) with red dotted box.}
\vspace{-0.8cm}
\end{figure*}

Different from the traditional way of extracting spatial features based on sliding window in CNN, the ConvLSTM cell holds three special gate mechanisms to complete data transmission and processing, which makes it possible to utilize spatial information of HSIs more effectively. Simultaneously, the ConvLSTM layer, constructed by the ConvLSTM cell in (3) as the basic unit, can not only implement data transmission and processing of the inter-layer, but also execute those of the intra-layer, which is another great difference from CNN. This special structure enables ConvLSTM layer to extract the more effective feature representation than CNN. Furthermore, compared with LSTM, the implementations of three gate mechanisms are extended from one-dimensional to multi-dimensional convolution operation, and this change can not only capture the spatial context information of the original data similar to the convolutional layer in CNN, but also model the long-range dependencies in the time domain of video or the spectral domain of HSIs.%, [53].

\section{Spatial-Spectral ConvLSTM 2-D Neural Network (SSCL2DNN)}
%Considering the shortcomings of LSTM, a modification and extended version of LSTM, i.e., ConvLSTM, is developed to build a new deep network for HSI feature extraction and classification. Depending on the different convolution structures in ConvLSTM, ConvLSTM2D with a 2-D convolution filter and ConvLSTM3D with a 3-D convolution filter can be obtained. These two extensions hold the same network structure with LSTM and can also be applied to model long-term dependencies in time domain and spectral domain.

In SSLSTMs \cite{FZ2017}, the spatial and spectral features are not effectively fused, and the spatial structure information is not well preserved. As for Bi-CLSTM \cite{QL2017}, the correlations between different spectral band is not fully utilized. In order to address the above issues, a novel ConvLSTM-based spatial-spectral feature extraction model is proposed. It should be noted that ConvLSTM given in the Section II-C is actually a 2-D extension structure of LSTM, and in what follows, to distinguish the 3-D extension form in Section VI, ConvLSTM involved in this Section is named ConvLSTM2D.

\subsection*{A. SSCL2DNN}

On account of the above analysis, a joint spatial-spectral feature extraction and classification model based on ConvLSTM2D is constructed, which is shown in Fig. 1 labeled by \textcircled{1} with blue. The core structure of our proposed SSCL2DNN model consists of ConvLSTM2D layer and pooling layer. Based on the basic ConvLSTM2D cell in (3), a ConvLSTM2D layer will be built, and a 3-D data cube needs to be decomposed into a spectral sequence, which is then input into each memory cell of the ConvLSTM2D layer one by one. However, there is much redundant information in original HSI data cubes, and the more data used, the greater complexity involved.
Unlike \cite{YC2016}, PCA is selected as the preprocessing method to implement dimension reduction of data.

First of all, the size of the original HSI data cube can be denoted as \(W \times H \times D\), where \(D\) indicates the number of the spectral bands, and \(W\) and \(H\) are the width and height of HSI, respectively. In order to reduce computational complexity, the first \( K \) components after PCA are selected as the spectral information of each pixel \(x_i\). Furthermore, for each ConvLSTM2D layer in SSCL2DNN, considering that the spatial context information is beneficial to HSI classification, the data in a local spatial window with the size of \(s \times s\) is extracted as the spatial information for extracting spatial-spectral features, which is the input of each memory cell in the ConvLSTM2D layer. After data preprocessing, a 3-D input data denoted by \(s \times s \times K\) is constructed. In particular, suppose that the special dimension time\_step in ConvLSTM2D is expressed as a variable \(\tau\), and it needs to be fixed as \( K \) to maintain the same dimension as the input data. Concretely, the 3-D input of each pixel \(x_i\) is decomposed into \(K\) 2-D components and converted into a sequence with the length of \( K \), i.e., \(\left\{X_i^1, \ldots, X_i^k, \ldots, X_i^K \right\}, k \in \left\{1,2,\ldots, K \right\}\), where \(X_i^k\) denotes the \(k\)th component of the pixel \(x_i\). Then, this sequence is fed into the ConvLSTM2D layer one by one.

In order to compress the feature maps generated by the ConvLSTM2D layer and reduce the computational complexity, the pooling layer is also used in the proposed deep model. After cascading multiple ConvLSTM2D layers and pooling layers, the final output of the last ConvLSTM2D layer in the proposed SSCL2DNN model is the desired feature representation that is subsequently fed into the classification layer to obtain the final result. Specifically, according to \cite{YC2016}, \cite{YL2017}, small convolution kernels are efficient for yielding better classification performance. As such, the size of the convolution kernels in our proposed deep model can be set as \(4 \times 4\) or \(3 \times 3\), and the kernel size of \(2 \times 2\) is adopted to implement the operation of the pooling layer.

Due to the exceptional inner structure of LSTM and its extended architecture, it is evident that if we set \(\tau\) to 1 and \( K \) to 1 in each ConvLSTM2D layer, and convert the input from a 3-D data to a 2-D data, the proposed deep model is reduced to spatial ConvLSTM 2-D Neural Network (SaCL2DNN), which is shown in Fig. 1 labeled by \textcircled{2} with the red dotted boxes. Specially, this transform is similar to that from 3D CNN to 2D CNN, however, SaCL2DNN can utilize the spatial structure information of hyperspectral data more effectively to obtain better classification performance than 2-D CNN.

Particularly, the proposed deep models can not only implement the same data transmission between layers as CNN does, but also accomplish effective data transmission and processing within each ConvLSTM2D layer. It is known that adjacent spectral bands are highly correlated, and there may also be some correlations between non-adjacent spectral bands \cite{HW2017}. Therefore, by effectively modeling long-range dependencies in the spectral field, SSCL2DNN can jointly consider the spatial-spectral information, which can address the problems of insufficient feature fusion caused by using two sub-branches independently and the loss of spatial information in SSLSTMs \cite{FZ2017} and underutilization of correlation between different spectral bands in Bi-CLSTM \cite{QL2017}.

%\textcolor[rgb]{0.00,0.00,1.00}{Compared with SSLSTMs \cite{FZ2017} and Bi-CLSTM \cite{QL2017}, SSCL2DNN can obtain more discriminative features by jointly considering the spatial-spectral information and modeling long-term dependencies in the spectral field, which can address the problem caused by using two sub-branches independently, and can settle over-fitting problem created by too many features and complexity to a certain extent.}

\subsection*{B. Loss Function and Optimization Method}
%\textcolor[rgb]{0.00,0.00,1.00}{By stacking multiple ConvLSTM2D layers and pooling layers in SSCL2DNN, there will be two kinds of outputs for the pixel \(x_i\) in the last ConvLSTM2D layer, i.e., \(\left\{H_i^1, \ldots, H_i^t, \ldots, H_i^K \right\}, t \in \left\{1,2,\ldots, K \right\}\) that the outputs of $K$ memory cells and \(\left\{C_{i}, H_{i}\right\}\) that the state and output generated by modeling long-range dependencies in the spectral field. Particularly, the output \(H_{i}\) is the desired spatial-spectral features in SSCL2DNN. After reducing dimensions through the last pooling layer, these features \(H_{i}\) are fed into a fully connected layer to map the feature space to class label space through linear transformation. Furthermore, the spatial-spectral features \(H_{i}\) of the pixel \(x_i\) in the proposed models are converted into 1-D vectors \(h_{i}\). Finally, these feature vectors are fed into the classification layer to predict the conditional probability distribution of each class as}
Suppose that there are $L$ ConvLSTM2D layers and pooling layers in SSCL2DNN. In the first ConvLSTM2D layer, each 2-D component \(X_i^k\) in the spectral sequence \(\left\{X_i^1, \ldots, X_i^k, \ldots, X_i^K \right\}\) is the input of the $k$th memory cell, which is actually \(X_t\) in (3), and then there will be two kinds of outputs, i.e., the outputs \(\left\{H_{1i}^1, \ldots, H_{1i}^k, \ldots, H_{1i}^K \right\}\) of $K$ memory cells and the state and output \(\left\{C_{1i}, H_{1i}\right\}\) yielded by modeling long-term dependencies, which are actually \(\left\{C_t, H_t\right\}\) in (3). The outputs of $K$ memory cells are retained as the inputs of the next layer to extract high-level features. After $L$ ConvLSTM2D layers and pooling layers, the output \( H_{Li}\) of the last ConvLSTM2D layer generated by modeling long-range dependencies in the spectral field is the desired spatial-spectral features. After reducing dimensions through the last pooling layer, the spatial-spectral features \( H_{Li}\) are converted into the 1-D vectors \(h_{Li}\) and then fed into a fully connected layer to map the feature space to class label space. Finally, the feature vectors \(h_{Li}\) are input to a softmax function to predict the conditional probability distribution $P(y=c|h_{Li},W,b)=\frac{e^{(W_{c}h_{Li}+b_c)}}{\sum_{j=1}^Ne^{(W_{j}h_{Li}+b_j)}}$ of each class $c$, where $c \in {1,2,\ldots,N}$, and $N$ is the number of classes in the HSI data sets.
%$P(y=c|h_{Li},W,b)= {}^{e^{(W_{c}h_{Li}+b_c)}}/_{\sum_{j=1}^Ne^{(W_{j}h_{Li}+b_j)}}$       ${}^1/_2$
%$P(y=c|h_{Li},W,b)=\frac{e^{(W_{c}h_{Li}+b_c)}}{\sum_{j=1}^Ne^{(W_{j}h_{Li}+b_j)}}$
%Finally, these feature vectors \(h_{Li}\) are input to the classification layer to predict the conditional probability distribution of each class as}
%\begin{equation}
%%\setlength{\abovedisplayskip}{4pt}
%%\setlength{\belowdisplayskip}{4pt}
%\begin{split}% ¹«Ê½×ó¶ÔÆë
%\textcolor[rgb]{0.00,0.00,1.00}{P_{c}=P(y=c|h_{Li},W,b)=\frac{e^{(W_{c}h_{Li}+b_c)}}{\sum_{j=1}^Ne^{(W_{j}h_{Li}+b_j)}}},
%\end{split}
%\end{equation}
%% Çó³öµÄ½á¹û±íʾµÄÊÇ Ã¿¸öÏñËصãxiÊôÓÚÀà±ðcµÄ¸ÅÂÊ¡£
%where \(W\) and \(b\) are the weight and bias of the classification layer, respectively. \textcolor[rgb]{0.00,0.00,1.00}{ $c$ denotes the class \(c\) in the HSI data sets, and \(c \in {1,2,\ldots,N}\).}%\(N\) is the number of classes in the HSI data sets.
\begin{figure*}[htbp]% ÍøÂçÄ£ÐÍ
\centering
%\vspace{-0.2cm}  %µ÷ÕûͼƬÓëÉÏÎĵĴ¹Ö±¾àÀë
\setlength{\abovecaptionskip}{-5pt}
\begin{center}
\includegraphics[height=1.65in, width=6.6in]{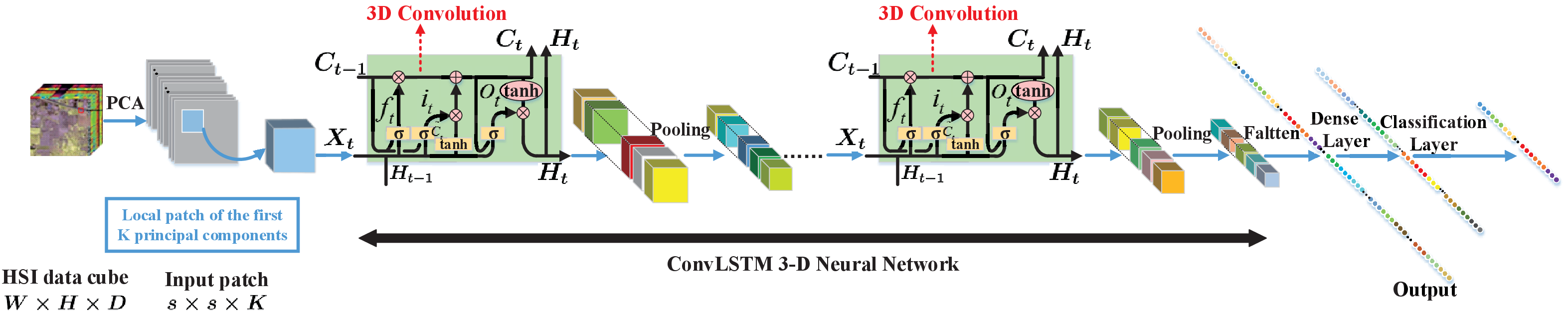}
%\caption{fig3}
\end{center}
%\tiny
\centering
\caption{The proposed spatial-spectral ConvLSTM 3-D Neural Network (SSCL3DNN) structure.}
\vspace{-0.4cm}
\end{figure*}

In addition, the cross entropy \cite{FZ2017} is used as the loss function to obtain the final classification results, which is optimized by adaptive momentum (ADAM) algorithm.% for the optimization of our loss function, the probability value \textcolor[rgb]{0.00,0.00,1.00}{\(P(y=c|x_{i}),c \in {1,2,\ldots,N}\)} of each pixel \(x_i\) can be obtained, where \(N\) indicates the number of classes in the HSI data sets..%, which can be described as
%\begin{equation}
%%\setlength{\abovedisplayskip}{4pt}
%%\setlength{\belowdisplayskip}{4pt}
%\begin{split}% ¹«Ê½×ó¶ÔÆë
%L_{loss}(Y,\tilde{Y})=-\sum(Y \cdot log(\tilde{Y})),
%\end{split}
%\end{equation}
%%\begin{equation}
%%%\setlength{\abovedisplayskip}{4pt}
%%%\setlength{\belowdisplayskip}{4pt}
%%\begin{split}% ¹«Ê½×ó¶ÔÆë
%%\textcolor[rgb]{0.00,0.00,1.00}{L_{loss}(Y,\tilde{Y})=-\sum_{c=1}^N(Y_{c} \cdot log(P_{c})),}
%%\end{split}
%%\end{equation}
%where \(Y\) and \textcolor[rgb]{0.00,0.00,1.00}{\(\tilde{Y}=\left\{P_{1},\ldots,P_{c},\ldots,P_{N}\right\}\)} are the ground truth in original HSI data sets and the corresponding predictive value of our deep \textcolor[rgb]{0.00,0.00,1.00}{model}, respectively.
% ´Ë´¦µÄÇóºÍÖ÷ÒªÊǶԵ±Ç°ÏñËصÄlabelÏòÁ¿ÓëÔ¤²âlabelÏòÁ¿µÄ¼ÆËã¡£

\section{Spatial-Spectral ConvLSTM 3-D Neural Network (SSCL3DNN)}

%Although SSCL2DNN may yield relatively good classification performance, the way of extracting spatial-spectral features by modeling long-range dependencies in spectral domain may not be the most effective approach, because the intrinsic structure of HSIs may be destroyed when taking each band component of the data patch as the input of each corresponding memory cell. Therefore, it may not be an elaborate spatial-spectral feature extraction method for HSI classification. This motivates us to further design a better deep network to capture the intrinsic structure of the hyperspectral data.
In order to better preserve the intrinsic structure information of the hyperspectral data, in this Section, the 3-D extended structure (ConvLSTM3D) is further developed, and another novel deep feature extraction and classification model, i.e., SSCL3DNN, is proposed, which can yield more discriminative spatial-spectral features for further improving the classification performance.

\subsection*{A. ConvLSTM3D}
On the basis of ConvLSTM2D, the 3-D extended version (ConvLSTM3D) is further developed, in which there are three gate units, and the calculation formulas are similar to that in ConvLSTM2D. However, different from it, the whole 3-D data cube is taken as the input of each memory cell in ConvLSTM3D. In particular, the input $\boldsymbol{\mathcal{X}_{t}}$, the state $\boldsymbol{\mathcal{C}_{t-1}}$ and $\boldsymbol{\mathcal{C}_{t}}$, the output $\boldsymbol{\mathcal{H}_{t-1}}$ and $\boldsymbol{\mathcal{H}_{t}}$, and the gate units $i_{t}$, $f_{t}$, and $o_{t}$ of ConvLSTM3D are 4-D tensors, whose last three dimensions are spectral dimension and two spatial dimensions, and the convolution filters \(\boldsymbol{\mathcal{W}_{\bullet i}}\), \(\boldsymbol{\mathcal{W}_{\bullet f}}\), and \(\boldsymbol{\mathcal{W}}_{\bullet o}\) are 3-D tensors. The structure illustration of the ConvLSTM3D layer is shown in Fig. 3. Specifically, the equations of the ConvLSTM3D cell can be written as
\begin{IEEEeqnarray}{lll}
i_t &=& \sigma(\boldsymbol{\mathcal{W}_{xi}} \circledast \boldsymbol{\mathcal{X}_{t}}+\boldsymbol{\mathcal{W}_{hi}} \circledast \boldsymbol{\mathcal{H}_{t-1}}+\boldsymbol{\mathcal{W}_{ci}} \circ \boldsymbol{\mathcal{C}_{t-1}}+b_i) \notag \\
f_t &=& \sigma({\small{\boldsymbol{\mathcal{W}_{xf}}}} \circledast \boldsymbol{\mathcal{X}_{t}}+ {\small{\boldsymbol{\mathcal{W}_{hf}}}} \circledast \boldsymbol{\mathcal{H}_{t-1}}+{\small{\boldsymbol{\mathcal{W}_{cf}}}} \circ \boldsymbol{\mathcal{C}_{t-1}}+b_f) \notag \\
\tilde{\boldsymbol{\mathcal{C}_{t}}} &=&\tanh(\boldsymbol{\mathcal{W}_{xc}} \circledast \boldsymbol{\mathcal{X}_{t}}+\boldsymbol{\mathcal{W}_{hc}} \circledast \boldsymbol{\mathcal{H}_{t-1}}+b_c)  \notag\\
\boldsymbol{\mathcal{C}_{t}} &=& f_t \circ \boldsymbol{\mathcal{C}_{t-1}} + i_t \circ \tilde{\boldsymbol{\mathcal{C}_{t}}} \notag \\
o_t &=& \sigma(\boldsymbol{\mathcal{W}_{xo}} \circledast \boldsymbol{\mathcal{X}_{t}}+\boldsymbol{\mathcal{W}_{ho}} \circledast \boldsymbol{\mathcal{H}_{t-1}}+\boldsymbol{\mathcal{W}_{co}} \circ \boldsymbol{\mathcal{C}_{t}}+b_o) \notag \\
\boldsymbol{\mathcal{H}_{t}} &=& o_t \circ \tanh(\boldsymbol{\mathcal{C}_{t}}).
\end{IEEEeqnarray}
where $\circledast$ is the defined 3-D convolution between 4-D input or output and 3-D convolution filter.

\begin{figure}[!htbp]% ÍøÂçÄ£ÐÍ
\centering
%\vspace{0.2cm}
\setlength{\abovecaptionskip}{-5pt}
\begin{center}
\includegraphics[width=3.1in]{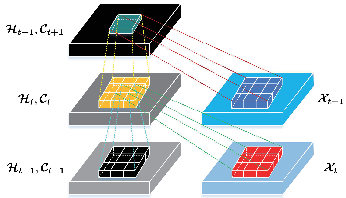}
\end{center}
\centering
\caption{The illustration of the inner structure of ConvLSTM3D.}
\vspace{-0.2cm}
\end{figure}
Compared with the LSTM cell in (2) and the ConvLSTM2D cell in (3), the extended ConvLSTM3D cell has a similar calculational model, however, the main difference is the calculation of the convolution in each gate unit. Taking the input gate of the ConvLSTM3D cell as an example. Suppose that $\boldsymbol{\mathcal{X}_{t}} \in R^{\tau_{t} \times w_{t} \times h_{t} \times s_{t}}$ is the input of it, which is the $t$th component in a sequence decomposed from the input of the ConvLSTM3D layer according to the dimension time\_step, and $\boldsymbol{\mathcal{W}_{xi}} \in R^{k_{1} \times k_{2} \times d}$, where $\tau_{t}$, $w_{t}$, $h_{t}$, $s_{t}$, $k_{1}$, $k_{2}$, and $d$ are the dimension time\_step, width, height, the number of the spectral bands, kernel size and depth, respectively. We define the 3-D convolution of $\boldsymbol{\mathcal{X}_{t}}$ and $\boldsymbol{\mathcal{W}_{xi}}$ as $\boldsymbol{\mathcal{W}_{xi}} \circledast \boldsymbol{\mathcal{X}_{t}}$, where the output $u^{\tau_{t}l_{x}l_{y}l_{z}}$ yielded by $\boldsymbol{\mathcal{X}_{t}}$ at position $(l_x,l_y,l_z)$ in the input gate is defined as
\begin{IEEEeqnarray}{lcl}
u^{\tau_{t}l_{x}l_{y}l_{z}} = \notag \\ \sum_{j=1}^{k_{1}}\sum_{p=1}^{k_{2}}\sum_{q=1}^{d}\boldsymbol{\mathcal{W}_{xi}}^{jpq}\boldsymbol{\mathcal{X}_{t}}^{\tau_{t} (l_x+j)(l_y+p)(l_z+q)}.% w,h,s ¿ÉÒÔÊÓΪÊÇ¿Õ¼äά¶ÈºÍ¹âÆ×ά¶È£¨ÒòΪÊÇÑØ×ÅÊäÈëµÄ¿Õ¼äά¶ÈºÍʱ¼äά¶È½øÐоí»ý¡££©
\end{IEEEeqnarray}

\begin{figure*}[htbp]
\centering
\vspace{-0.2cm}
\setlength{\abovecaptionskip}{-8pt}
\begin{center}
\includegraphics[width=6.5in]{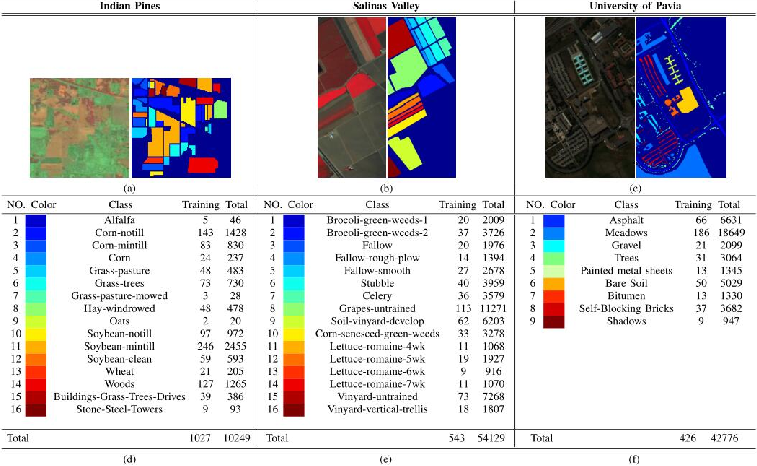}
%\caption{fig3}
\end{center}
%\tiny
\centering
\caption{(Left) False color maps and (Right) ground-truth maps of three HSI data sets. (a) Indian Pines (bands 20, 40, and 60). (b) Salinas Valley (bands 46, 27, and 10). (c) University of Pavia (bands 47, 27, and 13). (d)-(f) show the number of training samples. }
\vspace{-0.2cm}
\end{figure*}
\subsection*{B. SSCL3DNN}
According to the extraordinary structure of ConvLSTM3D, we further extend SSCL2DNN to a new deep model, i.e., SSCL3DNN, based on ConvLSTM3D. Different from SSCL2DNN, the whole data cube must be taken as the input of the memory cell in SSCL3DNN. Therefore, the intrinsic structure of HSIs can be captured well, and the spatial-spectral features can be well learned to further improve the HSI classification performance. Fig. 2 clearly shows the structure of the proposed deep model, and the parameter \(\tau\) is fixed as 1.

In particular, PCA is also used to reduce the redundant information in this Subsection, and similarly, the first \(K\) components are chosen. With regard to the selection of the size of the convolution kernel, the small kernel size is also considered, which can be set as \(3 \times 3 \times 3 \) or \(5 \times 5 \times 5 \) in the proposed deep model. Besides, the size of the pooling kernel in the pooling layer is set as \(2 \times 2 \times 2 \). After the cascade of multiple ConvLSTM3D layers and pooling layers, the spatial-spectral features are extracted from the 3-D data cube and subsequently fed into the classification layer to obtain the classification results.

Although the loss function and optimization method in this Section are the same as those of SSCL2DNN in Section III, the spatial-spectral features \(H_{Li}\) and its 1-D vector form \(h_{Li}\) of each pixel \(x_{i}\) and data processing at each layer in SSCL3DNN are quite different.

\section{Experimental Results}
To quantitatively and qualitatively analyze the classification performance of the proposed models, SVM \cite{CC2011}, 2-D CNN \cite{YC2016}, 3-D CNN \cite{YC2016}, SaLSTM \cite{FZ2017}, SSLSTMs \cite{FZ2017}, and Bi-CLSTM \cite{QL2017} are used as the comparative algorithms. In particular, the special case (SaCL2DNN) of SSCL2DNN is used to compare fairly with SVM, 2-D CNN, and SaLSTM. Three commonly used quantitative metrics are adopted, i.e., overall accuracy (OA), average accuracy (AA), and Kappa coefficient (\(\kappa\)). To eliminate the bias introduced by randomly choosing training samples, each experiment is repeated 10 times, and the mean values of each evaluation criterion are presented.

\subsection*{A. Hyperspectral Data Sets}
Three common HSI data sets, i.e., Indian Pines, Salinas Valley, and University of Pavia, are considered in our experiments, whose false color maps, groundtruth maps, and the corresponding training size are shown in Fig. 4, respectively.
%. There are the Indian Pines, Salinas Valley and University of Pavia (see Fig. 7). Specially, the first two data sets are obtained by the Airborne Visible/Infrared Imaging Spectrometer (AVIRIS) sensor, and the last one is acquired by the Reflective Optics System Imaging Spectrometer (ROSIS) sensor. Fig. 7(a), (b) and (c) apparently reveal the false color maps and ground-truth maps, and Fig. 7(d), (e) and (f) demonstrate the detailed data descriptions and the number of training samples of these three HSI data sets, respectively.

\emph{1) Indian Pines:} The Indian Pines data set was captured in 1992 by the Airborne Visible/Infrared Imaging Spectrometer (AVIRIS) sensor in Northwestern Indiana, USA, which is mainly composed of multiple agricultural fields. The spatial size of this data set is \(145 \times 145\) pixels with a spatial resolution of 20 meters per pixel (mpp), and there are 224 spectral bands in the wavelength range from 0.4 to 2.5 $\mu$m. Since some of them cannot be reflected by water together with four null bands, there are generally 200 bands remaining for study. After removing the background pixels, 10249 pixels are reserved, which contain useful ground-truth information from 16 different class labels.

\emph{2) Salinas Valley:} The Salinas Valley data set was collected by the 224-band AVIRIS sensor over Salinas Valley, California. This data set is made up of 512 lines and 217 columns, and contains 16 ground-truth classes. After removing the water absorption bands and noise-affected bands, there are 204 spectral bands preserved. %In addition, without considering the background pixels, there are 54129 pixels that consists of the ground-truth information in the form of a single label.

\emph{3) University of Pavia:} The University of Pavia data set was acquired by the Reflective Optics System Imaging Spectrometer (ROSIS) sensor over University of Pavia, Northern Italy. There are 103 spectral bands in the spectral range from 0.43 to 0.86 $\mu$m by removing several noise-corrupted bands, and it presents a size of \(610 \times 340\) pixels with a spatial resolution of 1.3 mpp. Different from the Indian Pines and Salinas Valley data sets, this data set contains 9 distinguishable classes. %Regardless of the background pixels, 42776 pixels are retained for research.

\subsection*{B. Experimental Settings}
In order to reduce the computational complexity, PCA is used as a preprocessing method to reduce data dimensions. For SVM, 2-D CNN, SaLSTM, and SaCL2DNN, the first principal component is preserved as the input. As far as SSCL2DNN and SSCL3DNN are concerned, the top \(K\) components are extracted as the spectral features.
%\textcolor[rgb]{0.00,0.00,1.00}{Since the first principal component in PCA contains the most valuable information of the hyperspectral data, PCA is selected to preserve only one component for SVM, 2-D CNN, SaLSTM, and SaCL2DNN. As for the proposed SSCL2DNN and SSCL3DNN models, after considering the redundancy and complexity of the hyperspectral data, PCA is also used to extract the top \(K\) principal components as the preserved spectral features.}

\begin{table}[H]% ·ÖÎö´°¿ÚÓëPCAÈ¡Öµ¶ÔÄ£ÐÍ·ÖÀàÐÔÄܵÄÓ°Ïì
    \centering
    \vspace{-0.1cm}  %µ÷ÕûͼƬÓëÉÏÎĵĴ¹Ö±¾àÀë
    \setlength{\abovecaptionskip}{-4pt}
    \renewcommand\thetable{\Roman{table}}
    \renewcommand\tabcolsep{1.0pt}
    \caption{Sensitivity Comparison and Analysis under Different Size ($s \times s$) of Local Windows.}
    %\footnotesize%\scriptsize
    \scriptsize%\scriptsize
    % ±íÍ·
    \begin{tabular}{p{2.72cm}<{\centering}p{2.0cm}<{\centering}p{2.0cm}<{\centering}}%{c|c|c}
    %\begin{tabular}{>{\columncolor{red}}p{3.9cm}<{\centering}>{\columncolor{red}}p{2.65cm}<{\centering}>{\columncolor{red}}p{4cm}<{\centering}}%{c|c|c}
%    	$ \begin{tabular}{c|c c|}
    	$ \begin{tabular}{c|c c|}
        \specialrule{0em}{6pt}{0pt}
        \hline
        \multicolumn{1}{c|}{\multirow{2}*{$s \times s$}}& \multicolumn{2}{c|}{Indian Pines} \\
%        \cline{2-6}
%        \specialrule{0em}{1pt}{0pt}
        \multicolumn{1}{c|}\quad& {\tiny SSCL2DNN} \quad& {\tiny SSCL3DNN} \\
        \hline
        %21$\times$21 & \multicolumn{1}{>{\columncolor{mycS14}}c}{96.63}         & 97.90         \\
        21$\times$21 & 96.63          & 97.90         \\
        23$\times$23 & 97.09          & 98.21          \\
        25$\times$25 & 97.37          & 98.32          \\
        \textbf{27$\times$27} & \textbf{97.72} & \textbf{98.79} \\
        29$\times$29 & 97.70          & 98.47          \\
        31$\times$31 & 97.68          & 98.20          \\
        33$\times$33 & 97.10          & 98.07          \\
        \hline
    \end{tabular}$ &
    	$ \begin{tabular}{c c c|}
        \specialrule{0em}{6pt}{0pt}
        \hline
        \multicolumn{3}{c|}{Salinas Valley} \\% ÿÀà10 ¸öµÄsalinas Êý¾ÝÓÐһЩÎÊÌâ
%        \cline{2-6}
%        \specialrule{0em}{1pt}{0pt}
        \quad& {\tiny SSCL2DNN} \quad& {\tiny SSCL3DNN}   \\
        \hline
        & 92.24          & 98.78          \\
        & 92.45          & 98.95          \\
        & 94.03          & 98.97          \\
        & \textbf{96.30} & 99.29          \\
        & 95.58          & 99.59          \\
        & 95.69          & \textbf{99.75} \\
        & 95.52          & 99.40          \\
        \hline
    \end{tabular}$ &
    	$ \begin{tabular}{c c c}
        \specialrule{0em}{6pt}{0pt}
        \hline
        \multicolumn{3}{c}{University of Pavia} \\
%        \cline{2-6}
%        \specialrule{0em}{1pt}{0pt}
        \quad& {\tiny SSCL2DNN} \quad& {\tiny SSCL3DNN}   \\
        \hline
        & 86.97          & 87.74          \\
        & 87.14          & 91.29          \\
        & 88.42          & 95.58          \\
        & \textbf{89.04} & 97.10          \\
        & 87.95          & 97.32          \\
        & 85.87          & \textbf{97.42} \\
        & 86.92          & 95.99          \\
        \hline
    \end{tabular}$ \\%\specialrule{0.05em}{0pt}{0.5pt}
    \end{tabular}
%\end{figure*}
\vspace{-0.9cm} %µ÷ÕûͼƬÓëÉÏÎĵĴ¹Ö±¾àÀë
\end{table}
\begin{table}[H]
    \centering
    \vspace{0.1cm}  %µ÷ÕûͼƬÓëÉÏÎĵĴ¹Ö±¾àÀë
    \setlength{\abovecaptionskip}{-4pt}
    \renewcommand\thetable{\Roman{table}}
    \renewcommand\tabcolsep{1.0pt}
    \caption{Sensitivity Comparison and Analysis under Different Number ($K$) of Principal Components After PCA.}
    %\footnotesize%\scriptsize
    \scriptsize
    % ±íÍ·
    \begin{tabular}{p{2.27cm}<{\centering}p{2.0cm}<{\centering}p{2.0cm}<{\centering}}%{c|c|c}
    	$ \begin{tabular}{c |c c|}
        \specialrule{0em}{6pt}{0pt}
        \hline
        \multicolumn{1}{c|}{\multirow{2}*{$K$}}& \multicolumn{2}{c|}{Indian Pines} \\% after Conducting PCA
        \multicolumn{1}{c|}\quad& {\tiny SSCL2DNN} \quad& {\tiny SSCL3DNN} \\ % after Conducting PCA
        \hline
        5  & \textbf{98.03} & 98.74 \\
        10 & 97.72          & \textbf{98.79} \\
        15 & 96.90          & 98.72          \\
        20 & 96.63          & 98.51          \\
        \hline
    \end{tabular}$ &
    	$ \begin{tabular}{c c c|}
        \specialrule{0em}{6pt}{0pt}
        \hline
        \multicolumn{3}{c|}{Salinas Valley} \\% ÿÀà10 ¸öµÄsalinas Êý¾ÝÓÐһЩÎÊÌâ
%        \cline{2-6}
%        \specialrule{0em}{1pt}{0pt}
        \quad& {\tiny SSCL2DNN} \quad& {\tiny SSCL3DNN} \\
        \hline
        & 95.64          & 99.07          \\
        & \textbf{96.30} & \textbf{99.29} \\
        & 95.24          & 99.18          \\
        &            &           \\
%        & 95.58          & 99.59          \\
        \hline
    \end{tabular}$ &
    	$ \begin{tabular}{c c c}
        \specialrule{0em}{6pt}{0pt}
        \hline
        \multicolumn{3}{c}{University of Pavia} \\
%        \cline{2-6}
%        \specialrule{0em}{1pt}{0pt}
        \quad& {\tiny SSCL2DNN} \quad& {\tiny SSCL3DNN} \\
        \hline
        & \textbf{91.47} & 93.43          \\
        & 89.04          & \textbf{97.10} \\
        & 85.56          & 96.74          \\
        & 83.46          & 96.53          \\
%        & 0          & 0          \\
        \hline
    \end{tabular}$ \\%\specialrule{0.05em}{0pt}{0.5pt}
    \end{tabular}
%\end{figure*}
\vspace{-0.6cm} %µ÷ÕûͼƬÓëÉÏÎĵĴ¹Ö±¾àÀë
\end{table}

%  ±í¸ñºóÃæ²»¼Ó[H]µÄ»°£¬±í¸ñλÖò»ÄÜÒƶ¯£¬Ö»»á·ÅÔÚÿһҳµÄ×ͷ
All the parameters of the compared methods are confirmed according to \cite{YC2016}, \cite{FZ2017}, \cite{QL2017}, \cite{CC2011} to achieve the quasi-optimal performance. Specifically, for SVM, the radial basis function (RBF) is used in the Libsvm toolbox \cite{CC2011}. There are two key parameters in SVM, i.e., \(C\) and \(\gamma\), which denote the regularization parameter and the kernel function parameter, respectively. According to \cite{CC2011}, fivefold cross-validation is adopted to tune \(C\) and \(\gamma\) from the range of \(\left\{2^{-5},2^{-4}, \ldots, 2^{19}\right\}\) and \(\left\{2^{-15},2^{-14}, \ldots, 2^{4} \right\}\), respectively. For Bi-CLSTM \cite{QL2017}, the kernel size and the number of feature maps in each ConvLSTM2D layer are fixed as 3 and 32, respectively. As for SeLSTM, SaLSTM, and SSLSTMs \cite{FZ2017}, the numbers of output nodes in SeLSTM and SaLSTM are set as 64 and 128, respectively, for the Indian Pines and Salinas Valley data sets, while as 128 and 256, respectively, for the University of Pavia data set.
\begin{table}[H]% ¡ª¡ª¡ª¡ª¡ª¡ª¡ª¡ª¡ª¡ª¡ª¡ª¡ª¡ª¡ª¡ª¡ª¡ª¡ª¡ª¡ª¡ªSaCL2DNN
    \centering
    \vspace{-0.1cm}  %µ÷ÕûͼƬÓëÉÏÎĵĴ¹Ö±¾àÀë
    \setlength{\abovecaptionskip}{-4pt}
    \renewcommand\thetable{\Roman{table}}
    \renewcommand\tabcolsep{3.0pt}
    \caption{The Parameter Settings of SaCL2DNN.}
    \label{tab:default}
    \scriptsize
    \begin{tabular}{p{2.12cm}<{\centering}|p{1.8cm}<{\centering}|p{1.5cm}<{\centering}|p{1.4cm}<{\centering}}%{c|c|c}
    %\begin{tabular}{p{2.2cm}<{\centering}|p{1.8cm}<{\centering}|p{1.6cm}<{\centering}|p{1.2cm}<{\centering}}%{c|c|c}
        \specialrule{0em}{6pt}{0pt}
        \specialrule{0.05em}{0pt}{0pt}
    	$ \begin{tabular}{c}
            %\textbf{Number} & \textbf{Color} & \textbf{Class} & \textbf{Training} & \textbf{Total}\\
            \specialrule{0em}{1pt}{1pt}
            \textbf{Layer Name} \\\specialrule{0em}{0pt}{1pt}
            \end{tabular}$ &
    	$ \begin{tabular}{c}
            \specialrule{0em}{1pt}{1pt}
            \textbf{Output Shape} \\\specialrule{0em}{0pt}{1pt}
            %\textbf{Number} & \textbf{Color} & \textbf{Class} & \textbf{Training} & \textbf{Total}\\
            \end{tabular}$ &
    	$ \begin{tabular}{c}
            \specialrule{0em}{1pt}{1pt}
            \textbf{Filter Size}\\\specialrule{0em}{0pt}{1pt}
            %\textbf{Number} & \textbf{Color} & \textbf{Class} & \textbf{Training} & \textbf{Total}\\
            \end{tabular}$ &
    	$ \begin{tabular}{c}
            \specialrule{0em}{1pt}{1pt}
            \textbf{Padding}\\\specialrule{0em}{0pt}{1pt}
            %\textbf{Number} & \textbf{Color} & \textbf{Class} & \textbf{Training} & \textbf{Total}\\
            \end{tabular}$ \\%\specialrule{0.05em}{0pt}{0.5pt}
    \end{tabular}
    % ÄÚÈÝ
    \begin{tabular}{p{2.2cm}<{\centering}|p{1.8cm}<{\centering}|p{1.5cm}<{\centering}|p{1.4cm}<{\centering}}%{c|c|c}
    %\begin{tabular}{p{2.2cm}<{\centering}|p{1.8cm}<{\centering}|p{1.6cm}<{\centering}|p{1.2cm}<{\centering}}%{c|c|c}
        \specialrule{0.05em}{0pt}{0pt}
    	$ \begin{tabular}{c}
            %\textbf{Number} & \textbf{Color} & \textbf{Class} & \textbf{Training} & \textbf{Total}\\
            \specialrule{0em}{1pt}{1pt}
            Input Layer \\
            ConvLSTM2D Layer \\
            MaxPooling Layer \\
            ConvLSTM2D Layer \\
            MaxPooling Layer \\
            Droup  \\
            Flatten \\
            Dense Layer \\
            %Droup  \\
            Output \\
            \end{tabular}$ &
    	$ \begin{tabular}{c}
            \specialrule{0em}{1pt}{1pt}
            \(27\times27\) \\
            \(27\times27\times32\) \\
            \(14\times14\times32\) \\
            \(14\times14\times64\) \\
            \(7\times7\times64\)\\
            \(7\times7\times64\) \\
            3136 \\
            128 \\
            %128  \\
            N \\
            \end{tabular}$ &
    	$ \begin{tabular}{c}
            \specialrule{0em}{1pt}{1pt}
             N/A \\
            \(3\times3\times32\)\\
            \(2\times2\times32\) \\
            \(5\times5\times64\) \\
            \(2\times2\times64\)\\
            0.25  \\
            N/A \\
            N/A \\
            %0.5  \\
            N \\
            \end{tabular}$ &
    	$ \begin{tabular}{c}
            \specialrule{0em}{1pt}{1pt}
             N/A \\
            SAME \\
            SAME \\
            SAME \\
            SAME\\
            N/A  \\
            N/A \\
            N/A \\
            %N/A  \\
            N/A \\
            \end{tabular}$ \\\specialrule{0.05em}{0pt}{0.5pt}
    \end{tabular}
    \vspace{-0.6cm}  %µ÷ÕûͼƬÓëÉÏÎĵĴ¹Ö±¾àÀë
\end{table}
\begin{table}[H]% ¡ª¡ª¡ª¡ª¡ª¡ª¡ª¡ª¡ª¡ª¡ª¡ª¡ª¡ª¡ª¡ª¡ª¡ª¡ª¡ª¡ª¡ªSSCL2DNN
    \centering
    \vspace{-0.1cm}  %µ÷ÕûͼƬÓëÉÏÎĵĴ¹Ö±¾àÀë
    \setlength{\abovecaptionskip}{-4pt}
    \renewcommand\thetable{\Roman{table}}
    \renewcommand\tabcolsep{3.0pt}
    \caption{The Parameter Settings of The Proposed SSCL2DNN.}
    \label{tab:default}
    \scriptsize
    \begin{tabular}{p{2.12cm}<{\centering}|p{2.3cm}<{\centering}|p{1.5cm}<{\centering}|p{1cm}<{\centering}}%{c|c|c}
    %\begin{tabular}{p{2.2cm}<{\centering}|p{2.3cm}<{\centering}|p{1.6cm}<{\centering}|p{1.2cm}<{\centering}}%{c|c|c}

        \specialrule{0em}{6pt}{0pt}
        \specialrule{0.05em}{0pt}{0pt}
    	$ \begin{tabular}{c}
            %\textbf{Number} & \textbf{Color} & \textbf{Class} & \textbf{Training} & \textbf{Total}\\
            \specialrule{0em}{1pt}{1pt}
            \textbf{Layer Name} \\\specialrule{0em}{0pt}{1pt}
            \end{tabular}$ &
    	$ \begin{tabular}{c}
            \specialrule{0em}{1pt}{1pt}
            \textbf{Output Shape} \\\specialrule{0em}{0pt}{1pt}
            %\textbf{Number} & \textbf{Color} & \textbf{Class} & \textbf{Training} & \textbf{Total}\\
            \end{tabular}$ &
    	$ \begin{tabular}{c}
            \specialrule{0em}{1pt}{1pt}
            \textbf{Filter Size}\\\specialrule{0em}{0pt}{1pt}
            %\textbf{Number} & \textbf{Color} & \textbf{Class} & \textbf{Training} & \textbf{Total}\\
            \end{tabular}$ &
    	$ \begin{tabular}{c}
            \specialrule{0em}{1pt}{1pt}
            \textbf{Padding}\\\specialrule{0em}{0pt}{1pt}
            %\textbf{Number} & \textbf{Color} & \textbf{Class} & \textbf{Training} & \textbf{Total}\\
            \end{tabular}$ \\%\specialrule{0.05em}{0pt}{0.5pt}
    \end{tabular}
    % ÄÚÈÝ
    \begin{tabular}{p{2.2cm}<{\centering}|p{2.3cm}<{\centering}|p{1.5cm}<{\centering}|p{1cm}<{\centering}}%{c|c|c}
    %\begin{tabular}{p{2.2cm}<{\centering}|p{2.3cm}<{\centering}|p{1.6cm}<{\centering}|p{1.2cm}<{\centering}}%{c|c|c}
        \specialrule{0.05em}{0pt}{0pt}
    	$ \begin{tabular}{c}
            %\textbf{Number} & \textbf{Color} & \textbf{Class} & \textbf{Training} & \textbf{Total}\\
            \specialrule{0em}{1pt}{1pt}
            Input Layer \\
            ConvLSTM2D Layer \\
            MaxPooling Layer \\
            ConvLSTM2D Layer \\
            MaxPooling Layer \\
            Droup  \\
            Flatten \\
            Dense Layer \\
            %Droup  \\
            Output \\
            \end{tabular}$ &
    	$ \begin{tabular}{c}
            \specialrule{0em}{1pt}{1pt}
            \(10\times27\times27\) \\
            \(10\times27\times27\times32\) \\
            \(10\times14\times14\times32\) \\
            \(1\times14\times14\times64\) \\
            \(1\times7\times7\times64\)\\
            \(1\times7\times7\times64\) \\
            3136 \\
            128 \\
            %128  \\
            N \\
            \end{tabular}$ &
    	$ \begin{tabular}{c}
            \specialrule{0em}{1pt}{1pt}
             N/A \\
            \(4\times4\times32\) \\
            \(2\times2\times32\) \\
            \(3\times3\times64\) \\
            \(2\times2\times64\)\\
            0.25  \\
            N/A \\
            N/A \\
            %0.5  \\
            N \\
            \end{tabular}$ &
    	$ \begin{tabular}{c}
            \specialrule{0em}{1pt}{1pt}
             N/A \\
            SAME \\
            SAME \\
            SAME \\
            SAME\\
            N/A  \\
            N/A \\
            N/A \\
            N/A \\
            \end{tabular}$ \\\specialrule{0.05em}{0pt}{0.5pt}
    \end{tabular}
    \vspace{-0.6cm}  %µ÷ÕûͼƬÓëÉÏÎĵĴ¹Ö±¾àÀë
\end{table}
\begin{table}[H]% ¡ª¡ª¡ª¡ª¡ª¡ª¡ª¡ª¡ª¡ª¡ª¡ª¡ª¡ª¡ª¡ª¡ª¡ª¡ª¡ª¡ª¡ªSSCL3DNN
    \centering
    \vspace{-0.1cm}  %µ÷ÕûͼƬÓëÉÏÎĵĴ¹Ö±¾àÀë
    \setlength{\abovecaptionskip}{-4pt}
    \renewcommand\thetable{\Roman{table}}
    \renewcommand\tabcolsep{3.0pt}
    \caption{The Parameter Settings of The Proposed SSCL3DNN.}
    \label{tab:default}
    \scriptsize
    \begin{tabular}{p{2.2cm}<{\centering}|p{2.2cm}<{\centering}|p{1.8cm}<{\centering}|p{1cm}<{\centering}}%{c|c|c}
        \specialrule{0em}{6pt}{0pt}
        \specialrule{0.05em}{0pt}{0pt}
    	$ \begin{tabular}{c}
            %\textbf{Number} & \textbf{Color} & \textbf{Class} & \textbf{Training} & \textbf{Total}\\
            \specialrule{0em}{1pt}{1pt}
            \textbf{Layer Name} \\\specialrule{0em}{0pt}{1pt}
            \end{tabular}$ &
    	$ \begin{tabular}{c}
            \specialrule{0em}{1pt}{1pt}
            \textbf{Output Shape} \\\specialrule{0em}{0pt}{1pt}
            %\textbf{Number} & \textbf{Color} & \textbf{Class} & \textbf{Training} & \textbf{Total}\\
            \end{tabular}$ &
    	$ \begin{tabular}{c}
            \specialrule{0em}{1pt}{1pt}
            \textbf{Filter Size}\\\specialrule{0em}{0pt}{1pt}
            %\textbf{Number} & \textbf{Color} & \textbf{Class} & \textbf{Training} & \textbf{Total}\\
            \end{tabular}$ &
    	$ \begin{tabular}{c}
            \specialrule{0em}{1pt}{1pt}
            \textbf{Padding}\\\specialrule{0em}{0pt}{1pt}
            %\textbf{Number} & \textbf{Color} & \textbf{Class} & \textbf{Training} & \textbf{Total}\\
            \end{tabular}$ \\%\specialrule{0.05em}{0pt}{0.5pt}
\end{tabular}
    % ÄÚÈÝ
\begin{tabular}{p{2.2cm}<{\centering}|p{2.2cm}<{\centering}|p{1.8cm}<{\centering}|p{1cm}<{\centering}}%{c|c|c}
        \specialrule{0.05em}{0pt}{0pt}
    	$ \begin{tabular}{c}
            %\textbf{Number} & \textbf{Color} & \textbf{Class} & \textbf{Training} & \textbf{Total}\\
            \specialrule{0em}{1pt}{1pt}
            Input Layer \\
            ConvLSTM3D Layer \\
            MaxPooling Layer \\
            ConvLSTM3D Layer \\
            MaxPooling Layer \\
            Droup  \\
            Flatten \\
            Dense Layer \\
            Droup  \\
            Output \\
            \end{tabular}$ &
    	$ \begin{tabular}{c}
            \specialrule{0em}{1pt}{1pt}
            \(10\times27\times27\) \\
            \(10\times27\times27\times32\) \\
            \(5\times14\times14\times32\) \\
            \(5\times14\times14\times64\) \\
            \(3\times7\times7\times64\)\\
            \(3\times7\times7\times64\) \\
            9408 \\
            128 \\
            128  \\
            N \\
            \end{tabular}$ &
    	$ \begin{tabular}{c}
            \specialrule{0em}{1pt}{1pt}
             N/A \\
            \(4\times4\times4\times32\) \\
            \(2\times2\times2\times32\) \\
            \(3\times3\times3\times64\) \\
            \(2\times2\times2\times64\)\\
            0.25  \\
            N/A \\
            N/A \\
            0.5  \\
            N \\
            \end{tabular}$ &
    	$ \begin{tabular}{c}
            \specialrule{0em}{1pt}{1pt}
             N/A \\
            SAME \\
            SAME \\
            SAME \\
            SAME\\
            N/A  \\
            N/A \\
            N/A \\
            N/A  \\
            N/A \\
            \end{tabular}$ \\\specialrule{0.05em}{0pt}{0.5pt}
\end{tabular}
\vspace{-0.4cm}  %µ÷ÕûͼƬÓëÉÏÎĵĴ¹Ö±¾àÀë
\end{table}
%===============================================Section C µÄ·ÂÕæʵÑé2 2D HSI classification and 3D HSI classification  8 ÖÖ·½·¨
\begin{table*}% Indian 10%
    \centering
    \vspace{-0.2cm}  %µ÷ÕûͼƬÓëÉÏÎĵĴ¹Ö±¾àÀë
    \setlength{\abovecaptionskip}{-4pt}
    \renewcommand\thetable{\Roman{table}}
    \renewcommand\tabcolsep{3.0pt}
    \caption{Classification Results for The Indian Pines Data Set Using 10\% Training Samples.}
    \scriptsize%
    \begin{tabular}{c|c c c c | c c c c c}
        %\hline\noalign{\smallskip}
        \specialrule{0em}{6pt}{0pt}
        \hline
        Class \quad& SVM \quad& 2-D CNN \quad& SaLSTM \quad& SaCL2DNN \quad& 3-D CNN \quad& SSLSTMs \quad& Bi-CLSTM \quad& SSCL2DNN \quad& SSCL3DNN \\
        \hline
        1 & 70.73  & \textbf{90.24}& 85.98 & 86.59         & 99.39          & 79.88 & 91.06         & \textbf{100.00}& 98.17\\
        2 & 89.32  & 95.47         & 92.43 & \textbf{96.73}& 97.86          & 94.24 & 94.29         & 98.11 & \textbf{99.12}\\
        3 & 91.93  & \textbf{95.21}& 87.25 & 94.31         & \textbf{97.05} & 90.70 & 93.13         & 96.56 & 96.52\\
        4 & 86.50  & 92.84         & 88.97 & \textbf{93.54}& 96.95          & 88.85 & 88.89         & 96.56 & \textbf{98.71}\\
        5 & 90.57  & \textbf{94.02}& 91.49 & 91.55         & 96.55          & 91.09 & 94.25         & 96.09 & \textbf{97.07}\\
        6 & 97.72  & 97.45         & 95.81 & \textbf{98.36}& 99.20          & 96.58 & \textbf{99.49}& 98.02 & 99.09\\
        7 & 58.00  & \textbf{75.00}& 67.00 & 74.00         & 94.00          & 77.00 & 93.33         & 84.00 & \textbf{96.00}\\
        8 & 98.72  & 99.07         & 96.16 & \textbf{99.71}& \textbf{100.00}& 97.09 & 99.46         & 99.69 & \textbf{100.00}\\
        9 & 44.44  & 61.11         & 59.72 & \textbf{63.89}& \textbf{76.39} & 63.89 & 38.89         & 55.56 & 63.89\\
        10 & 77.20 & 94.86         & 89.37 & \textbf{95.54}& 97.31          & 91.49 & 95.73         & 97.07 & \textbf{98.74}\\
        11 & 95.02 & 98.03         & 95.30 & \textbf{99.33}& 98.95          & 95.21 & 96.60         & \textbf{99.34} & 99.33\\
        12 & 84.27 & 91.06         & 85.53 & \textbf{95.27}& 97.94          & 86.42 & 89.37         & 96.75 & \textbf{98.78}\\
        13 & 89.13 & 93.75         & 86.68 & \textbf{97.96}& 97.01          & 91.98 & 95.65         & 98.19 & \textbf{99.73}\\
        14 & 97.41 & 98.79         & 97.72 & \textbf{99.34}& 99.56          & 98.51 & \textbf{99.80}& 99.65 & 99.69\\
        15 & 93.73 & 97.33         & 95.24 & \textbf{98.70}& 99.06          & 94.42 & 97.12         & 98.85 & \textbf{99.42}\\
        16 & 68.75 & \textbf{93.15}& 70.83 & 88.39         & \textbf{94.35} & 79.46 & 89.29         & 85.32 & 92.56\\
        \hline
        OA & 91.20         & 96.21 & 92.59 & \textbf{97.07} & 98.28          & 93.69 & 95.62 & 98.03 & \textbf{98.79}\\
        AA & 83.34         & 91.71 & 86.59 & \textbf{92.08} & \textbf{96.35} & 88.56 & 90.90 & 93.73 & 96.05\\
        \(\kappa\) & 89.91 & 95.67 & 91.53 & \textbf{96.66} & 98.04          & 92.79 & 94.99 & 97.75 & \textbf{98.62}\\
        \hline
    \end{tabular}
    \vspace{-0.35cm}  %µ÷ÕûͼƬÓëÉÏÎĵĴ¹Ö±¾àÀë
\end{table*}
\begin{table*}% % Salinas 1%
\centering
%\vspace{-0.2cm}  %µ÷ÕûͼƬÓëÉÏÎĵĴ¹Ö±¾àÀë
\setlength{\abovecaptionskip}{-4pt}
\renewcommand\thetable{\Roman{table}}
\renewcommand\tabcolsep{3.0pt}
\caption{Classification Results for The Salinas Valley Data Set Using 1\% Training Samples.}
\scriptsize%\scriptsize
\begin{tabular}{c|c c c c | c c c c c}
    %\hline\noalign{\smallskip}
    \specialrule{0em}{6pt}{0pt}
    \hline
    Class \quad& SVM \quad& 2-D CNN \quad& SaLSTM \quad& SaCL2DNN \quad& 3-D CNN \quad& SSLSTMs \quad& Bi-CLSTM \quad& SSCL2DNN \quad& SSCL3DNN \\
    \hline
    1  & 74.26 & 81.35          & 89.66 & \textbf{96.68} & 98.59          & 82.02 & 93.08 & 91.64 & \textbf{99.73}\\
    2  & 85.23 & 86.21          & 76.85 & \textbf{97.40} & 99.96          & 82.15 & 99.57 & 97.80 & \textbf{100.00}\\
    3  & 59.56 & 78.14          & 51.94 & \textbf{81.61} & 96.22          & 54.62 & 98.99 & 98.77 & \textbf{100.00}\\
    4  & 98.19 & 97.66          & 95.97 & \textbf{99.13} & 99.81          & 95.53 & 98.84 & 99.40 & \textbf{100.00}\\
    5  & 94.33 & 93.45          & 93.66 & \textbf{97.62} & \textbf{99.66} & 93.05 & 99.41 & 98.83 & 99.14\\
    6  & 94.40 & 94.61          & 94.01 & \textbf{99.63} & 99.98          & 95.74 & 99.76 & \textbf{100.00}& \textbf{100.00}\\
    7  & 78.97 & 88.90          & 80.20 & \textbf{95.48} & 98.74          & 82.30 & 98.15 & 96.23 & \textbf{100.00}\\
    8  & 85.36 & 93.66          & 82.68 & \textbf{93.52} & 89.17          & 85.36 & 90.05 & 93.05 & \textbf{97.36}\\
    9  & 68.44 & 84.14          & 80.87 & \textbf{95.91} & 99.84          & 77.75 & 99.69 & 99.89 & \textbf{99.97}\\
    10 & 71.65 & 87.81          & 70.49 & \textbf{95.78} & 99.04          & 70.77 & 99.62 & 98.44 & \textbf{99.76}\\
    11 & 93.98 & 94.54          & 97.67 & \textbf{96.53} & 97.45          & 95.90 & 99.12 & 98.08 & \textbf{99.68}\\
    12 & 88.38 & 87.47          & 84.22 & \textbf{95.25} & 99.09          & 86.60 & 97.43 & 99.06 & \textbf{100.00}\\
    13 & 80.04 & 82.21          & 72.14 & \textbf{86.99} & 97.46          & 68.72 & 99.23 & 97.02 & \textbf{99.45}\\
    14 & 84.04 & 88.10          & 92.73 & \textbf{93.08} & 99.84          & 91.34 & \textbf{99.87} & 99.69 & 99.75\\
    15 & 90.36 & 94.59          & 86.30 & \textbf{98.11} & \textbf{99.49} & 88.93 & 89.46 & 92.78 & 99.48\\
    16 & 64.92 & \textbf{78.96} & 64.82 & 71.16          & 99.85          & 61.13 & 97.34 & 93.39 & \textbf{100.00}\\
    \hline
    OA & 82.39         & 89.65 & 82.28 & \textbf{94.68} & 97.17 & 83.10 & 95.72 & 96.30 & \textbf{99.29}\\
    AA & 82.01         & 88.24 & 82.14 & \textbf{93.37} & 98.39 & 81.99 & 97.48 & 97.13 & \textbf{99.65}\\
    \(\kappa\) & 80.30 & 88.47 & 80.26 & \textbf{94.06} & 96.85 & 81.15 & 95.24 & 95.88 & \textbf{99.21}\\
    \hline
\end{tabular}
\vspace{-0.35cm}  %µ÷ÕûͼƬÓëÉÏÎĵĴ¹Ö±¾àÀë
\end{table*}
\begin{table*}% Pavia 1%
    \centering
    %\vspace{-0.4cm}  %µ÷ÕûͼƬÓëÉÏÎĵĴ¹Ö±¾àÀë
    \setlength{\abovecaptionskip}{-4pt}
    \renewcommand\thetable{\Roman{table}}
    \renewcommand\tabcolsep{3.0pt}
    \caption{Classification Results for The University of Pavia Data Set Using 1\% Training Samples.} % OF THE AVAILABLE LABELED DATA FOR TRAINING AND \(27 \times 27\) INPUT LOCAL PATCH SIZE (AVERAGE OF \(10\) RUNS)} \(\pm\) STANDARD DEVIATION)}
    \scriptsize%\footnotesize
    \begin{tabular}{c|c c c c | c c c c c}
        %\hline\noalign{\smallskip}
        \specialrule{0em}{6pt}{0pt}
        \hline
        Class \quad& SVM \quad& 2-D CNN \quad& SaLSTM \quad& SaCL2DNN \quad& 3-D CNN \quad& SSLSTMs \quad& Bi-CLSTM \quad& SSCL2DNN \quad& SSCL3DNN \\
        \hline
        1 & 52.21 & 79.64         & 57.22 & \textbf{80.20} & 82.91 & 70.57 & 88.24          & 96.15 & \textbf{97.62}\\
        2 & 89.78 & 93.79         & 81.84 & \textbf{98.54} & 96.25 & 89.90 & 98.18          & 99.43 & \textbf{99.95}\\
        3 & 17.73 & 44.26         & 27.74 & \textbf{45.76} & 66.25 & 32.20 & 37.91          & 64.28 & \textbf{82.96}\\
        4 & 55.23 & 77.30         & 57.98 & \textbf{81.89} & 90.61 & 69.21 & 80.39          & 90.45 & \textbf{96.52}\\
        5 & 68.05 & 77.76         & 70.80 & \textbf{79.55} & 93.83 & 77.69 & 80.56          & 89.01 & \textbf{98.51}\\
        6 & 39.14 & 70.75         & 42.07 & \textbf{85.55} & 85.72 & 50.47 & 66.77          & 76.60 & \textbf{97.75}\\
        7 & 15.19 & \textbf{38.22}& 27.46 & 36.05          & 77.80 & 35.20 & 56.09          & 67.15 & \textbf{88.61}\\
        8 & 60.56 & 84.19         & 62.86 & \textbf{87.23} & 88.48 & 68.19 & 82.22          & 92.58 & \textbf{95.41}\\
        9 & 53.39 & 79.13         & 57.23 & \textbf{90.06} & 68.64 & 85.37 & \textbf{86.67} & 78.04 & 83.60\\
        \hline
        OA & 65.67         & 81.89 & 64.77 & \textbf{86.69} & 89.14 & 73.90 & 85.22 & 91.47 & \textbf{97.10}\\
        AA & 50.14         & 71.67 & 53.91 & \textbf{76.09} & 83.39 & 64.31 & 75.22 & 83.74 & \textbf{93.44}\\
        \(\kappa\) & 51.52 & 75.86 & 52.90 & \textbf{81.94} & 85.59 & 64.42 & 79.85 & 88.49 & \textbf{96.14}\\
        \hline
    \end{tabular}
    \vspace{-0.85cm}
\end{table*}
\begin{table*}  % ÿÀà10¸ö  2D  ûÓбê×¼²î
    \centering
    \vspace{-0.2cm}  %µ÷ÕûͼƬÓëÉÏÎĵĴ¹Ö±¾àÀë
    \setlength{\abovecaptionskip}{-4pt}
    \renewcommand\tabcolsep{1.0pt}
    \caption{Classification Results of Training under Small Samples (10 Samples for Each Class).}
    %LASSIFICATION RESULTS OF TRAINING UNDER SMALL SAMPLES (10 SAMPLES FOR EACH CLASS)} %AND \(27 \times 27\) INPUT LOCAL PATCH SIZE (AVERAGE OF \(10\) RUNS)} \(\pm\) STANDARD DEVIATION)}
    \scriptsize%\footnotesize%
    % ±íÍ·
    \begin{tabular}{p{4.7cm}<{\centering}p{3.7cm}<{\centering}p{5cm}<{\centering}}%{c|c|c}
    	$ \begin{tabular}{c|c c c c|}
        \specialrule{0em}{6pt}{0pt}
        \hline
        \multicolumn{1}{c|}{\multirow{2}*{Class}}& \multicolumn{4}{c|}{Indian Pines} \\
%        \cline{2-6}
%        \specialrule{0em}{1pt}{0pt}
        \multicolumn{1}{c|}\quad& SVM \quad& \(2\)-D CNN \quad& SaLSTM & \multicolumn{1}{c|}{SaCL2DNN}\\
        \hline
        1  & 91.67          & 94.44          & 88.33 & \textbf{95.00}\\
        2  & \textbf{42.99} & 37.83          & 31.20 & 38.98\\
        3  & \textbf{56.51} & 47.59          & 41.90 & 52.34\\
        4  & 73.83          & 76.12          & 65.37 & \textbf{84.93}\\
        5  & 56.87          & 58.01          & 57.67 & \textbf{63.00}\\
        6  & \textbf{60.42} & 46.06          & 43.86 & 51.44\\
        7  & 97.78          & \textbf{100.00}& 94.44 & \textbf{100.00}\\
        8  & \textbf{96.28} & 93.33          & 70.56 & 87.78\\
        9  & \textbf{100.00}& \textbf{100.00}& 84.00 & \textbf{100.00}\\
        10 & 52.77          & 47.80          & 42.64 & \textbf{56.30}\\
        11 & \textbf{42.94} & 42.45          & 28.09 & 42.81\\
        12 & 49.67          & 45.97          & 38.42 & \textbf{52.08}\\
        13 & 81.13          & 86.97          & 75.18 & \textbf{95.90}\\
        14 & 70.66          & 69.94          & 73.34 & \textbf{82.10}\\
        15 & 71.33          & 69.95          & 67.98 & \textbf{78.67}\\
        16 & 90.84          & 90.60          & 77.11 & \textbf{92.77}\\
        \hline
        OA & 56.41 & 53.08 & 45.77 & \textbf{57.47}\\
        AA & 70.98 & 69.19 & 61.26 & \textbf{73.38}\\
        \(\kappa\) & 51.34 & 47.72 & 40.40 & \textbf{52.75}\\
        \hline
    \end{tabular}$ &
    	$ \begin{tabular}{c c c c c|}
        \specialrule{0em}{6pt}{0pt}
        \hline
        \multicolumn{5}{c|}{Salinas Valley} \\% ÿÀà10¸öµÄsalinasÊý¾ÝÓÐһЩÎÊÌâ
%        \cline{2-6}
%        \specialrule{0em}{1pt}{0pt}
        \quad& SVM \quad& \(2\)-D CNN \quad& SaLSTM & \multicolumn{1}{c|}{SaCL2DNN}\\
        \hline
        &73.31          & 76.16          & 70.96 & \textbf{83.71}\\
        &58.15          & 61.93          & 47.37 & \textbf{73.62}\\
        &53.38          & 46.20          & 36.31 & \textbf{67.70}\\
        &95.94          & 97.34          & 93.05 & \textbf{99.21}\\
        &87.47          & 88.78          & 82.36 & \textbf{92.50}\\
        &\textbf{87.51} & 85.21          & 81.16 & 82.60\\
        &63.57          & 68.06          & 60.57 & \textbf{82.64}\\
        &37.40          & \textbf{38.37} & 40.94 & 32.38\\
        &43.82          & 34.70          & 40.19 & \textbf{45.79}\\
        &53.64          & 53.20          & 47.65 & \textbf{83.46}\\
        &94.67          & 93.01          & 89.21 & \textbf{96.07}\\
        &89.17          & 80.86          & 69.13 & \textbf{90.28}\\
        &90.53          & 93.09          & 85.76 & \textbf{94.88}\\
        &90.60          & 87.72          & 88.72 & \textbf{97.96}\\
        &\textbf{81.05} & 78.31          & 77.32 & 72.90\\
        &73.32          & 61.37          & 51.50 & \textbf{82.53}\\
        \hline
        &63.77 & 62.15 & 59.15 & \textbf{67.49}\\
        &73.28 & 71.52 & 66.39 & \textbf{79.89}\\
        &60.24 & 58.56 & 55.22 & \textbf{64.47}\\
        \hline
    \end{tabular}$ &
    	$ \begin{tabular}{c c c c c}
        \specialrule{0em}{6pt}{0pt}
        \hline
        \multicolumn{5}{c}{University of Pavia} \\
%        \cline{2-6}
%        \specialrule{0em}{1pt}{0pt}
        \quad& SVM \quad& \(2\)-D CNN \quad& SaLSTM \quad& SaCL2DNN \\
        \hline
        &40.07          & 37.14          & 32.95 & \textbf{44.89}\\
        &42.49          & 43.51          & 30.42 & \textbf{67.08}\\
        &30.80          & \textbf{34.96} & 21.97 & 34.67\\
        &54.21          & 59.74          & 31.83 & \textbf{66.72}\\
        &\textbf{83.97} & 75.18          & 69.46 & 80.02\\
        &33.57          & 23.30          & 18.84 & \textbf{45.57}\\
        &\textbf{66.39} & 35.86          & 33.23 & 39.77\\
        &43.75          & 63.47          & 30.15 & \textbf{79.58}\\
        &57.42          & 86.27          & 57.45 & \textbf{95.55}\\
        &  &  &  &   \\
        &  &  &  &   \\
        &  &  &  &   \\
        &  &  &  &   \\
        &  &  &  &   \\
        &  &  &  &   \\
        &  &  &  &   \\
        \hline
        & 43.81 & 44.30 & 31.02 & \textbf{60.76}\\
        & 50.30 & 51.05 & 36.26 & \textbf{61.54}\\
        & 32.36 & 32.59 & 19.04 & \textbf{50.60}\\
        \hline
    \end{tabular}$ \\%\specialrule{0.05em}{0pt}{0.5pt}
    \end{tabular}
%\end{figure*}
\vspace{-0.1cm}  %µ÷ÕûͼƬÓëÉÏÎĵĴ¹Ö±¾àÀë
\end{table*}
\begin{figure*}[htbp]
\centering
%\vspace{-0.2cm}
\setlength{\abovecaptionskip}{-8pt}
\begin{center}
\includegraphics[width=5.8in]{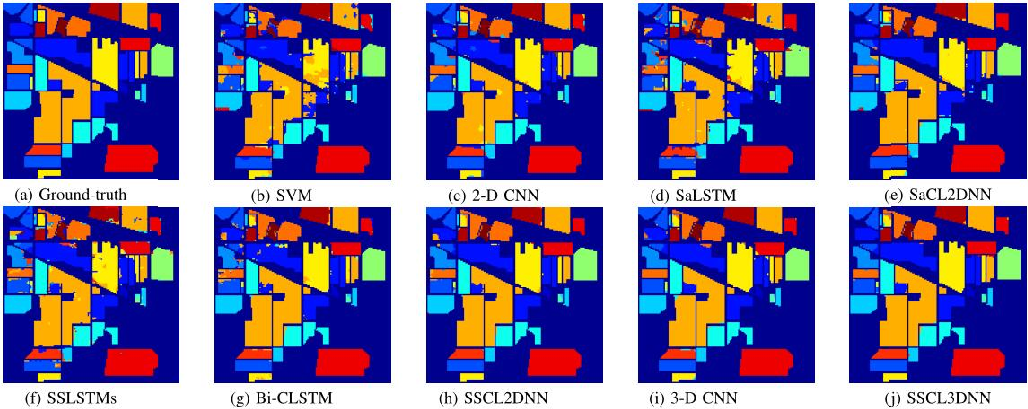}
%\caption{fig3}
\end{center}
%\tiny
\centering
\caption{Classification maps for the Indian Pines data set. (a) Ground-truth map. (b) SVM. (c) 2-D CNN. (d) SaLSTM. (e) SaCL2DNN. (f) SSLSTMs. (g) Bi-CLSTM. (h) SSCL2DNN. (i) 3-D CNN. (j) SSCL3DNN.}
\vspace{-0.2cm}
\end{figure*}
For SSCL2DNN and SSCL3DNN, there are four key parameters to be determined, i.e., the size (\(s \times s\)) of the local window, the number (\(K\)) of the principal components after PCA, the size (\(k \times k\)) of the convolution operation, and the number ($M$) of the feature maps at each ConvLSTM layer. Firstly, \(K\) is fixed as 10. Then, $s$ is generated from $\left\{{21, 23, 25, 27, 29, 31, 33}\right\}$, $k$ is from $\left\{{3, 4, 5}\right\}$, and $M$ is searched from three given combinations $\left\{{16,32}\right\}$, $\left\{{32,64}\right\}$, and $\left\{{64,128}\right\}$. The experimental results under different size of local window are reported in Table I, from which it can be seen that %with the increase of the value of $s$, the classification accuracy of these two models increases first and then decreases. Specifically,
for these three data sets, the optimal \(s\) in SSCL2DNN is 27, while for SSCL3DNN, it is 27, 31, and 31, respectively. Although SSCL3DNN can obtain 0.46\% and 0.32\% gains when \(s\) varies from 27 to 31 for the Salinas Valley and University of Pavia data sets, the computational complexity and runtime will obviously increase. Based on the above analysis, the size of local window for the proposed models is fixed as \(27 \times 27\), which can not only achieve satisfactory classification performance, but also provide convenience for the practical application of the proposed models. In particular, the size of local window in SaCL2DNN is the same as that in SSCL2DNN.

After that, the experiments for analyzing the influence of different \(K\) on the performance are further completed, and $K$ is searched from $\left\{{5, 10, 15, 20}\right\}$ for the Indian Pines and University of Pavia data sets, while from $\left\{{5, 10, 15}\right\}$ for the Salinas Valley data set because of the memory problems. The results in Table II reveal that for these three data sets, the optimal $K$ in SSCL3DNN is 10, while for SSCL2DNN, it is 5, 10, and 5, respectively.

In order to ensure the convergence of the loss function, the number of training epochs is fixed as 2000, and the learning rate is set as 0.0001 from epochs 1 to 2000 for the proposed deep models. More detailed parameter settings are summarized in Tables III-V.

For hardware system configuration, all the following experiments are completed on a desktop with an 8th Generation Intel Core i7-8700 processor and up to 3.7 GHz, 16 GB of DDR4 RAM with a serial speed of 2400 MHz, a Nvidia GeForce GTX 1080ti GPU with 11 GB memory, an Inter SSD D3-S4510 with 240GB. For software system configuration, we adopt Ubuntu 16.04.4 x64 as our operating system for all experiments. CUDA 8.0 and cuDNN 7.0.5, Tflearn with 0.3.2, Tensorflow-gpu with 1.4.0 and python 3.5.4 are the main programming environment. Specially, all methods involved in our experiments are completed in Anaconda 3.4.

It should be noted that Tflearn is a modular and transparent deep learning library built on Tensorflow, and can provide a higher level API than Tensorflow. A combination of Tflearn and Tensorflow constructs the core framework of the proposed deep models.

\subsection*{C. Classification Performance}

\begin{table*}  % ÿÀà10¸ö 3D ûÓбê×¼²î
    \centering
    %\vspace{-0.25cm}  %µ÷ÕûͼƬÓëÉÏÎĵĴ¹Ö±¾àÀë
    \setlength{\abovecaptionskip}{-4pt}
    \renewcommand\tabcolsep{0.5pt}
    \caption{Classification Results of Training under Small Samples (10 Samples for Each Class).}
    \scriptsize%\footnotesize
    % ±íÍ·
    \begin{tabular}{p{5cm}<{\centering}p{4.2cm}<{\centering}p{5cm}<{\centering}}
        $ \begin{tabular}{c|c c c c c|}
            \specialrule{0em}{6pt}{0pt}
            \hline
            \multicolumn{1}{c|}{\multirow{2}*{Class}}& \multicolumn{5}{c|}{Indian Pines} \\
            \multicolumn{1}{c|}\quad& {\tiny 3-D CNN} \quad& {\tiny SSLSTMs} \quad& {\tiny Bi-CLSTM} \quad& {\tiny SSCL2DNN} & \multicolumn{1}{c|}{{\tiny SSCL3DNN}}\\
            \hline
            1  & 97.22          & 86.11 & 98.15          & 98.15          & \textbf{100.00}\\
            2  & \textbf{53.60} & 35.35 & 27.41          & 38.65          & 53.30\\
            3  & \textbf{61.78} & 38.44 & 39.02          & 46.34          & 50.37\\
            4  & \textbf{96.83} & 62.03 & 81.20          & 79.88          & 88.63\\
            5  & 75.90          & 61.27 & 66.67          & 61.95          & \textbf{76.58}\\
            6  & 78.47          & 40.39 & 74.77          & 65.56          & \textbf{90.17}\\
            7  & \textbf{100.00}& 97.78 & \textbf{100.00}& \textbf{100.00}& \textbf{100.00}\\
            8  & 99.83          & 71.67 & 97.79          & 94.59          & \textbf{100.00}\\
            9  & \textbf{100.00}& 92.00 & \textbf{100.00}& \textbf{100.00}& \textbf{100.00}\\
            10 & 63.56          & 44.86 & 53.40          & 66.22          & \textbf{75.36}\\
            11 & 55.09          & 32.74 & \textbf{61.83} & 50.46          & 59.66\\
            12 & \textbf{67.03} & 42.98 & 58.43          & 60.83          & 64.97\\
            13 & 96.82          & 82.87 & 93.16          & 96.58          & \textbf{99.79}\\
            14 & 88.11          & 71.81 & 83.16          & 73.86          & \textbf{88.29}\\
            15 & 84.20          & 66.60 & 82.98          & 84.13          & \textbf{87.66}\\
            16 & 98.31          & 86.99 & 90.36          & 91.16          & \textbf{100.00}\\
            \hline
            OA & 69.21         & 47.57 & 61.91 & 60.59 & \textbf{71.28}\\
            AA & 82.30         & 63.37 & 75.52 & 75.52 & \textbf{83.42}\\
            \(\kappa\) & 65.56 & 42.16 & 57.33 & 56.12 & \textbf{67.83}\\
            \hline
        \end{tabular}$ &
        $ \begin{tabular}{c c c c c c|}
            \specialrule{0em}{6pt}{0pt}
            \hline
            \multicolumn{6}{c|}{Salinas Valley} \\
            \quad& {\tiny 3-D CNN} \quad& {\tiny SSLSTMs} \quad& {\tiny Bi-CLSTM} \quad& {\tiny SSCL2DNN} & \multicolumn{1}{c|}{{\tiny SSCL3DNN}} \\
            \hline
            & \textbf{99.84}& 74.81 & 74.95         & 79.13 & 99.69\\
            & 89.36         & 50.32 & 83.88         & 70.94 & \textbf{99.90}\\
            & 92.56         & 39.67 & 98.07         & 98.80 & \textbf{99.93}\\
            & 99.36         & 94.86 & 99.33         & 98.22 & \textbf{99.83}\\
            & 96.25         & 83.22 & \textbf{98.50}& 94.80 & 98.01\\
            & 99.22         & 90.47 & 95.71         & 94.02 & \textbf{99.67}\\
            & 95.39         & 66.74 & 90.59         & 87.12 & \textbf{99.74}\\
            & \textbf{87.12}& 44.20 & 64.87         & 65.67 & 85.00\\
            & 96.33         & 40.61 & 99.25         & 97.44 & \textbf{99.73}\\
            & 95.18         & 47.55 & 94.72         & 91.98 & \textbf{99.04}\\
            & 97.05         & 89.62 & 98.20         & 98.03 & \textbf{99.41}\\
            & 97.27         & 71.95 & 97.76         & 97.00 & \textbf{99.69}\\
            & 98.87         & 85.52 & \textbf{99.67}& 98.92 & 99.65\\
            & 98.49         & 87.13 & 99.09         & 97.00 & \textbf{99.75}\\
            & 63.49         & 77.58 & 64.95         & 71.52 & \textbf{86.22}\\
            & 95.72         & 51.60 & 88.17         & 79.59 & \textbf{99.98}\\
            \hline
            & 89.77 & 61.63 & 83.88 & 82.92 & \textbf{94.73}\\
            & 93.84 & 68.49 & 90.48 & 88.76 & \textbf{97.83}\\
            & 88.57 & 57.92 & 82.12 & 81.05 & \textbf{94.14}\\
            \hline
        \end{tabular}$ &
        $ \begin{tabular}{c c c c c c}
            \specialrule{0em}{6pt}{0pt}
            \hline
            \multicolumn{5}{c}{University of Pavia} \\
            \quad& {\tiny 3-D CNN} \quad& {\tiny SSLSTMs} \quad& {\tiny Bi-CLSTM} \quad& {\tiny SSCL2DNN} & \multicolumn{1}{c}{{\tiny SSCL3DNN}} \\
            \hline
            & 42.58          & 39.94 & 51.32 & 49.43 & \textbf{69.89}\\
            & \textbf{84.89} & 29.08 & 82.59 & 76.89 & 82.44\\
            & 25.50          & 25.10 & 33.89 & 39.38 & \textbf{65.10}\\
            & 34.01          & 60.82 & 52.32 & 62.48 & \textbf{90.48}\\
            & 91.41          & 73.71 & 86.52 & 84.42 & \textbf{94.83}\\
            & 37.47          & 29.14 & 45.37 & 41.29 & \textbf{82.37}\\
            & 80.61          & 40.66 & 70.96 & 61.16 & \textbf{89.32}\\
            & 10.74          & 34.70 & 57.75 & 66.68 & \textbf{84.71}\\
            & 21.10          & 85.91 & 84.95 & 83.46 & \textbf{86.91}\\
            &  &  &  &  & \\
            &  &  &  &  & \\
            &  &  &  &  & \\
            &  &  &  &  & \\
            &  &  &  &  & \\
            &  &  &  &  & \\
            &  &  &  &  & \\
            \hline
            & 58.50 & 36.33 & 66.49 & 64.59 & \textbf{81.11}\\
            & 47.59 & 46.56 & 62.85 & 62.80 & \textbf{82.90}\\
            & 44.68 & 25.33 & 55.57 & 54.45 & \textbf{75.65}\\
            \hline
    \end{tabular}$ \\
    \end{tabular}
%\end{figure*}
\vspace{-0.32cm}  %µ÷ÕûͼƬÓëÉÏÎĵĴ¹Ö±¾àÀë
\end{table*}
\begin{figure*}[htbp]
\centering
%\vspace{-0.2cm}
\setlength{\abovecaptionskip}{-8pt}
\begin{center}
\includegraphics[width=5.5in]{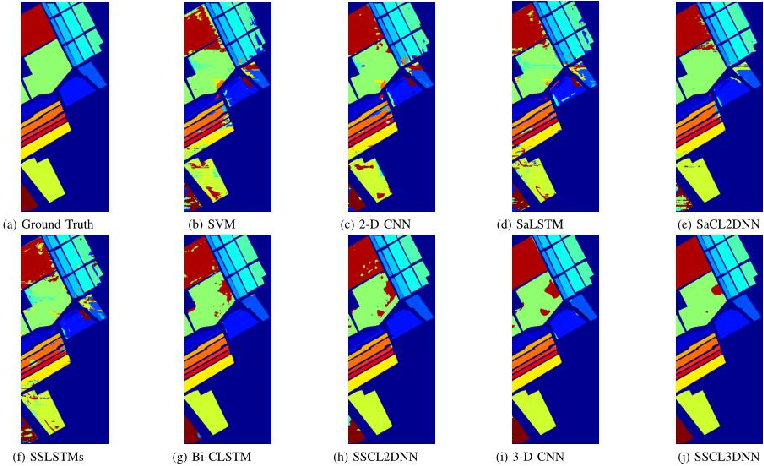}
%\caption{fig3}
\end{center}
%\tiny
\centering
\caption{Classification maps for the Salinas Valley data set. (a) Ground-truth map. (b) SVM. (c) 2-D CNN. (d) SaLSTM. (e) SaCL2DNN. (f) SSLSTMs. (g) Bi-CLSTM. (h) SSCL2DNN. (i) 3-D CNN. (j) SSCL3DNN.}
\vspace{-0.2cm}
\end{figure*}

In order to show the superiority of the proposed deep models, we randomly select \(1\%\) of the available labeled samples for training for both the Salinas Valley and University of Pavia data sets while \(10\%\) for the Indian Pines data set, respectively. And the remaining samples are used for testing.

According to the experiment settings in Subsection B, the quantitative assessments based on all models are reported in Tables VI-VIII, from which it can be seen that the proposed SSCL2DNN and SSCL3DNN models can provide better classification performance than other considered models. First of all, three special gate mechanisms make it possible for ConvLSTM to take full use of both the spatial and spectral information of HSIs than CNN that is operated by the traditional method based on the sliding window. What's more, the implementations of the gate mechanisms in ConvLSTM are extended from one-dimensional to multi-dimensional convolution operation, which enables ConvLSTM to better preserve and capture spatial context information of HSIs, and fuse the spatial and spectral information more effectively than LSTM. Specifically, on the one hand, compared with 2-D CNN, SaCL2DNN improves OA by 0.86\%, 5.03\%, and 4.80\%, respectively, for the Indian Pines, Salinas Valley, and University of Pavia data sets, which can better capture spatial context information from hyperspectral data. It should be noted that it is the loss of the spatial information that leads to unsatisfactory performance for SVM and SaLSTM. In addition, SaCL2DNN is a special case of SSCL2DNN, and compared with it, SSCL2DNN can obtain 0.96\%, 1.62\%, and 4.78\% gains in OA for these three data sets, respectively. For Bi-CLSTM, although ConvLSTM2D is also the core backbone for the feature extraction, this way of simply cascading all the outputs not only does not fully exploit the correlations between different spectral bands, but also easily leads to overfitting problem. Different from it, SSCL2DNN is built by alternating cascades of multi-layer ConvLSTM2D layers and pooling layers, in which the outputs yielded by modeling long-range dependencies are the final feature representations. To some extent, this approach can not only make great use of the characteristics of ConvLSTM2D, but also reduce the number of features and the complexity of the model, and the gains in OA yielded by SSCL2DNN are 2.41\%, 0.58\%, and 6.25\% for these three HSI data sets, respectively, which verifies the effectiveness of SSCL2DNN, and demonstrates that joint learning of spatial-spectral features by modeling long-term dependencies in the spectral field can provide higher classification performance.

%============================================Section E µÄ·ÂÕæʵÑé ͼ10(OA ÇúÏßͼ) £¨Ð¡Ñù±¾·ÖÎö£©
On the other hand, SSCL3DNN fuses the spatial and spectral information more effectively by the special 3-D operation, which can preserve the intrinsic structure of hyperspectral data to further improve the classification performance of SSCL2DNN, and obtains 0.76\%, 2.99\%, and 5.63\% gains in OA for these three HSI data sets, respectively. Compared with 3-D CNN, the improvements in OA provided by SSCL3DNN are 0.51\%, 2.12\%, and 7.96\% for three data sets, respectively. In particular, SSCL3DNN produces the remarkable gains for the Salinas Valley and University of Pavia data sets. From Tables VI-VIII, it can be seen that the 3-D extended architecture of LSTM makes it possible for SSCL3DNN to generate better classification performance by preserving the intrinsic structure of hyperspectral data, and the special gate structures enable SSCL3DNN to extract more discriminative spatial-spectral features. In addition, the classification performance of some classes with high correlation, such as class 10, class 11, and class 12 in the Indian Pines data set, class 8 and class 15 in the Salinas Valley data set, and class 1, class 6, and class 8 in the University of Pavia data set, is improved, which demonstrates the superiority of the proposed models.
\begin{figure*}[htbp]
\centering
%\vspace{-0.2cm}
\setlength{\abovecaptionskip}{-8pt}
\begin{center}
\includegraphics[width=5.5in]{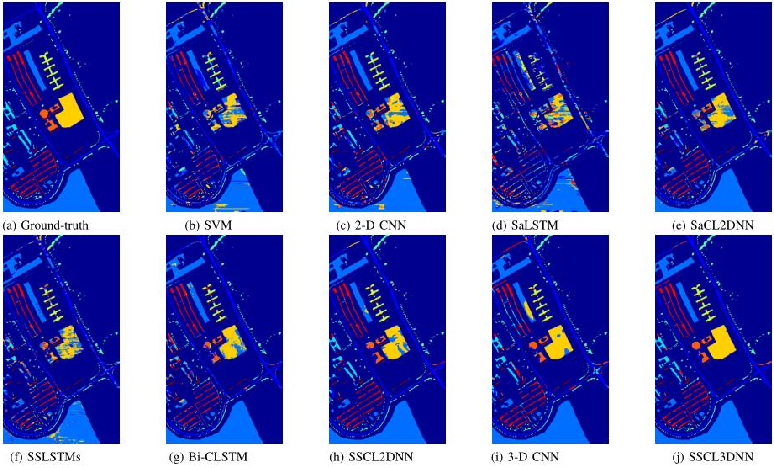}
%\caption{fig3}
\end{center}
%\tiny
\centering
\caption{Classification maps for the University of Pavia data set. (a) Ground-truth map. (b) SVM. (c) 2-D CNN. (d) SaLSTM. (e) SaCL2DNN. (f) SSLSTMs. (g) Bi-CLSTM. (h) SSCL2DNN. (i) 3-D CNN. (j) SSCL3DNN.}
\vspace{-0.2cm}
\end{figure*}

Corresponding to Tables VI-VIII, similar conclusions can be drawn from the classification maps presented in Figs. 5-7, from which it is obvious that the maps provided by the proposed models are closest to the ground-truth maps for these three data sets. In addition, there are only fewer misclassifications, and the boundaries of each class are better recognized, especially for class 2, class 10, and class 11 in Fig. 5, class 8 and class 15 in Fig. 6, and class 1, class 6, and class 8 in Fig. 7, which further verifies the effectiveness of the proposed models.

As we all know, it is greatly expensive and difficult to obtain samples with labels. Therefore, it is necessary to investigate the performance under small training size. %More detailed sensitivity comparison and analysis under small training samples will be given in Section D.
\begin{figure*}[htbp]
%\begin{table*}
    \centering
    \renewcommand\tabcolsep{1.0pt}
    \scriptsize
    \begin{tabular}{p{5.65cm}<{\centering}p{5cm}<{\centering}p{6cm}<{\centering}}
        \begin{minipage}[t]{1\linewidth}
            \centering
            \includegraphics[height=2.1in, width=2.1in]{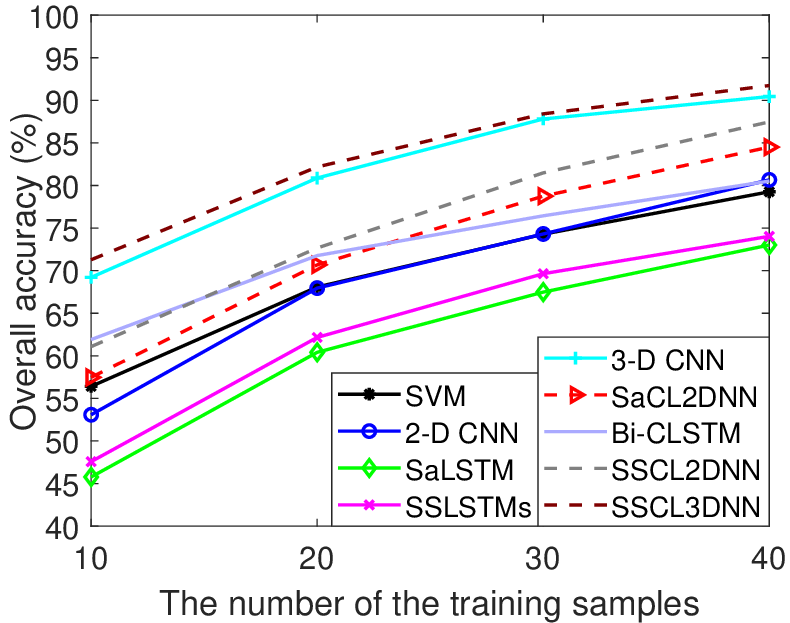}
        \end{minipage}
        & \begin{minipage}[t]{1\linewidth}
            \centering
            \includegraphics[height=2.1in, width=2.1in]{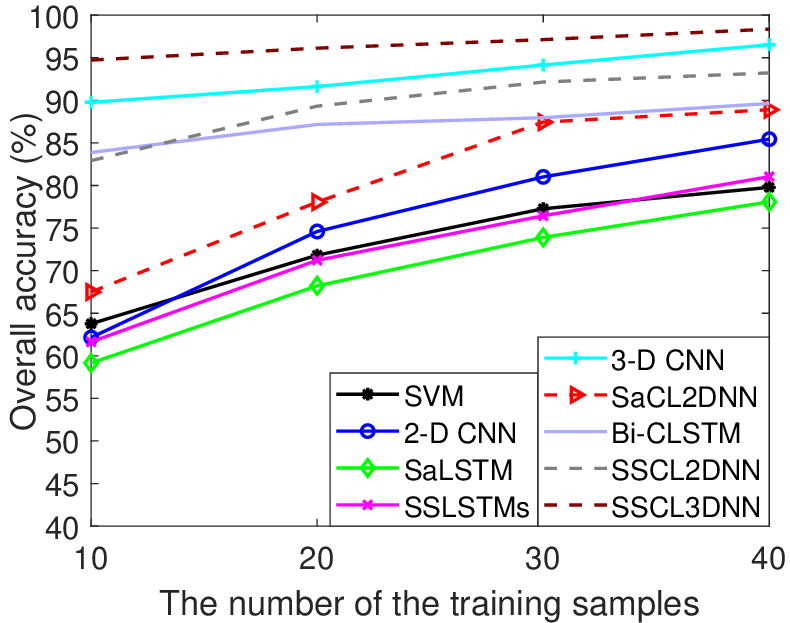}
        \end{minipage}
        & \begin{minipage}[t]{1\linewidth}
            \centering
            \includegraphics[height=2.1in, width=2.1in]{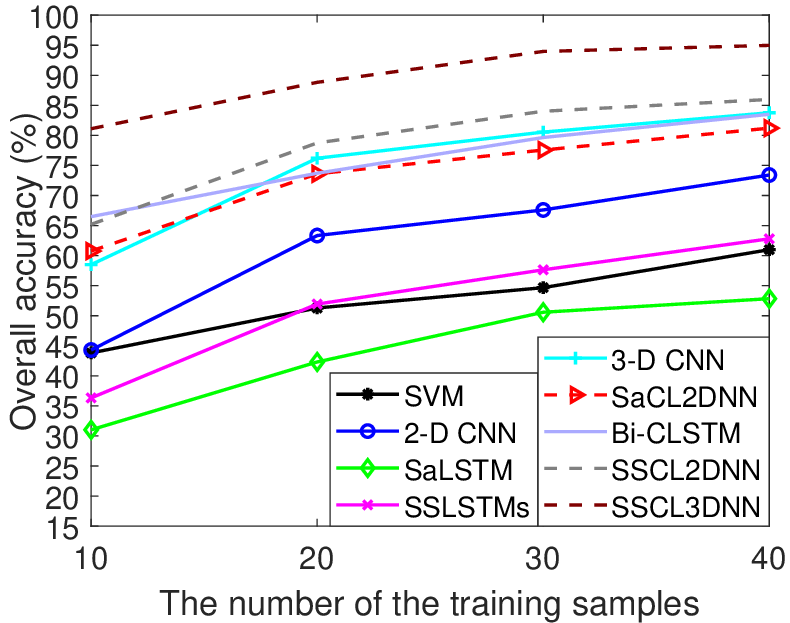}
        \end{minipage}\\
        %\specialrule{0em}{1pt}{1pt}
        (a) Indian Pines& (b) Salinas Valley & (c) University of Pavia \\
    \end{tabular}
\caption{Overall accuracy (\%) of considered methods with different number of training samples for three HSI data sets: (a) Indian Pines, (b) Salinas Valley, (c) University of Pavia.}
\vspace{-0.4cm}  %µ÷ÕûͼƬÓëÉÏÎĵĴ¹Ö±¾àÀë
\end{figure*}

\subsection*{D. Sensitivity Comparison and Analysis under Small Samples}

In order to further demonstrate the performance of the proposed deep models, we randomly select 10 samples from each labeled class to construct smaller training sets.

The experimental results are reported in Tables IX-X, from which it can be observed that even in the case of small size of training samples, the proposed deep models can also show better classification performance. Compared with 2-D CNN, the improvements in OA yielded by SaCL2DNN are 4.39\%, 5.34\%, and 16.46\%, respectively, and SSCL2DNN improves the performance of SaCL2DNN and obtains 3.12\%, 15.43\%, and 3.83\% gains in OA for these three HSI data sets, respectively, which further verifies the validity of joint learning of the spatial-spectral features. However, Bi-CLSTM obtains better performance than SSCL2DNN, and yields 1.32\%, 0.96\%, and 1.90\% gains in OA, respectively. The main reason is the way of cascading all the outputs of each ConvLSTM2D layer determines that Bi-CLSTM can provide more available features than SSCL2DNN. In particular, it is the defects of Bi-CLSTM and SSCL2DNN that lead to the destruction of the intrinsic structure of the hyperspectral data when putting each component of the local patch as the input of each memory cell. SSCL3DNN overcomes this shortcomings, and yields the best classification performance, which obtains 2.07\%, 4.96\%, and 22.61\% gains in OA for these three data sets, respectively, when compared with 3-D CNN. More detailed results are reported in Tables IX-X, which further demonstrate the advantages of the proposed models.

To further show the effectiveness of the proposed deep models, 20, 30, and 40 samples from each class are randomly extracted for training in these three data sets. In particular, the number of training samples for class 7 and class 9 in the Indian Pines data set is fixed as 10. When taking different numbers of training samples into account, the OA curves of all models are provided in Fig. 8, from which it can be seen that for these three HSI data sets, SSCL3DNN provides the highest classification accuracy. It is worth noting that SSCL3DNN obtains significant performance improvements in the Salinas Valley and University of Pavia data sets. In particular, compared with Bi-CLSTM, SSCL2DNN achieves higher accuracy when the number of training samples is greater than or equal to 20, which means the way of using ConvLSTM2D to model long-range dependencies between different spectral bands may be more effective than that of simply cascading all the outputs. The results in Tables IX-X and Fig. 8 further demonstrate the effectiveness of the proposed deep models, and the design of the special gate structure makes it possible for them to better capture spatial information and preserve the intrinsic structure information of the original data, which is extremely important for improving the classification performance.

\section{Conclusion}
In this paper, two novel deep ConvLSTM Neural networks, i.e., SSCL2DNN and SSCL3DNN, have been proposed to extract more effective and discriminative spatial-spectral features for HSI classification. In SSCL2DNN, by taking the local patch as a spectral sequence and feeding into each memory cell band by band, the outputs by modeling long-range dependencies in the spectral domain are the spatial-spectral features, which can reduce the number of features to solve the overfitting problem to a certain extent while achieving satisfactory classification performance. By further developing the 3-D extended structure of LSTM, SSCL3DNN can better preserve the intrinsic structure of hyperspectral data, and the special gate mechanisms enable it to extract more discriminative spatial-spectral features, which can further improve classification performance. The experimental results conducted on three widely used HSI data sets show that the proposed deep models offer competitive advantages over state-of-the-art approaches, especially in the case of small training size.

\section*{Acknowledgment}
The authors would like to thank the Associate Editor and the Anonymous Reviewers for their valuable comments and suggestions, which are greatly helpful to improve the quality and presentation of this paper.
% or


%%%\appendices
%%%\section{Proof of the First Zonklar Equation}
%%%Appendix one text goes here.
%%%
%%%% you can choose not to have a title for an appendix
%%%% if you want by leaving the argument blank
%%%\section{}
%%%Appendix two text goes here.

%\section*{ACKNOWLEDGMENT}

\ifCLASSOPTIONcaptionsoff
  \newpage
\fi

\end{spacing}
\end{document}